\documentclass[conference]{IEEEtran}
\usepackage{hyperref}
\usepackage{url}

\usepackage{nomencl}
\makenomenclature

\usepackage{graphicx}
\graphicspath{ {Figures/} }

\usepackage{amsmath,bm}
\usepackage{amssymb}

\usepackage{cite}
\usepackage{amsmath}
\usepackage{algpseudocode}

\usepackage{array}

\ifCLASSOPTIONcompsoc
  \usepackage[caption=false,font=normalsize,labelfont=sf,textfont=sf]{subfig}
\else
  \usepackage[caption=false,font=footnotesize]{subfig}
\fi

\usepackage{stfloats}
\usepackage{url}

\begin{document}
%
\title{Deep Neural Network Based Subspace Learning of Robotic Manipulator Workspace Mapping}

\nomenclature[04]{$\fontfamily{phv}\selectfont \textbf{A}$}{A tensor}%
\nomenclature[06]{$\textbf{A}^T$}{Transpose of matrix $\textbf{A}$}%
\nomenclature[07]{det$(\textbf{A})$}{Determinant of $\textbf{A}$}%


\nomenclature[11]{$p($\textbf{x}$)$}{A probability distribution over a continuous variable}%
\nomenclature[12]{x $\sim p$}{Random variable $x$ has distribution $p$}%
\nomenclature[13]{$\mathbb{E}_{x \sim p}[f(x)]$}{Expectation of $f(x)$ with respect to $P(x)$}%
\nomenclature[14]{$\mathcal{N}(\textbf{\textit{x}};\mu,\sigma)$}{Gaussian distribution over $\textit{\textbf{x}}$ with mean $\mu$ and
covariance $\sigma$ }%

\nomenclature[15]{$\begin{Bmatrix} a,b,c \end{Bmatrix}$ or $\mathbb{F}$}{A set}%
\nomenclature[19]{$\mathbb{B}$}{Boolean domain}%
\nomenclature[21]{$SO(n)$}{Special orthogonal group, the set of all orientations in $n$ dimensions}%
\nomenclature[22]{$\textbf{\textit{R}}$}{Orthogonal rotation matrix, $\textbf{\textit{R}} \in SO(3)$}%
\nomenclature[23]{$\mathcal{G}$}{A Graph}%
\nomenclature[24]{$Pa_{\mathcal{G}}(x_i)$}{The parents of $x_i$ in $\mathcal{G}$}%

\nomenclature[25]{$f(\textbf{\textit{x}};\theta)$}{A function of $\textbf{\textit{x}}$ parametrized by $\theta$}%
\nomenclature[26]{$\textbf{H}(f)(\textbf{\textit{x}})$}{The Hessian matrix of $f$ at input point $\textbf{\textit{x}}$}%
\nomenclature[27]{$\xi$}{Abstract representation of 3-dimensional Cartesian pose}%
\nomenclature[28]{$\xi (\textbf{\textit{a}},\textbf{\textit{R}})$}{Pose construction function that takes in a position vector and an orientation matrix and returns a pose in $SE(3)$}%
\nomenclature[29]{$\mathcal{K}(\textbf{q})$}{Forward kinematic function of a serial-link robot}%
\nomenclature[30]{$\mathcal{K}^{-1}(\xi)$}{Inverse kinematics}%
\nomenclature[32]{$\textbf{\textit{x}}^{(i)}$}{The $i$-th example from a dataset}%
\nomenclature[33]{$\textbf{\textit{y}}^{(i)}$}{The target associated with $\textbf{\textit{x}}^{(i)}$ for supervised learning}%
\nomenclature[34]{$||\textbf{\textit{x}}||$}{$L^2$ norm of $\textbf{\textit{x}}$}%
\nomenclature[35]{$\delta($x$;\mu)$ or $\delta(\mu)$}{Dirac delta distribution of a scalar random variable centered at $\mu$}%
\nomenclature[36]{$c$}{Constant with significance indicated by context}%
\nomenclature[37]{$\mathcal{U}(a_{min},a_{max})$}{Uniform distribution of a random variable on the interval $[a_{min},a_{max}]$}%

\author{\IEEEauthorblockN{Peiyuan Liao}
\IEEEauthorblockA{
Kent School\\
Kent, Connecticut 06757\\
Email: liaop20@kent-school.edu}
}


%


\maketitle

\begin{abstract}
The manipulator workspace mapping is an important problem in robotics and has attracted significant attention in the community. However, most of the pre-existing
 algorithms have expensive time complexity due to the reliance on sophisticated kinematic equations. To solve this problem, this paper introduces subspace learning (SL), 
a variant of subspace embedding, where a set of robot and scope parameters is mapped to the corresponding workspace by a deep neural network (DNN). 
Trained on a large dataset of around $\mathbf{6\times 10^4}$ samples obtained from a MATLAB\textsuperscript \textregistered \space implementation of a classical method and sampling of designed uniform distributions, 
the experiments demonstrate that the embedding significantly reduces run-time from $\mathbf{5.23 \times 10^3}$ s of traditional discretization method to 
$\mathbf{0.224}$ s, with high accuracies (average F-measure is $\mathbf{0.9665}$ with batch gradient descent and resilient backpropagation).\footnote{MATLAB\textsuperscript \textregistered \space Scripts are available at \url{https://github.com/liaopeiyuan/Subspace-Learning}}
\end{abstract}


%
\IEEEpeerreviewmaketitle

\section{Introduction}
As an essential feature of robotic manipulators, manipulator workspace is the set of poses they can attain by their end effectors. Since the last few decades, the its study has gained much attention by the field because the workspace of a manipulator foreshadows its functionality and understanding it aids in the process of trajectory planning. Thus, extensive research has been done on developing an algorithm to calculate the workspace \cite{AW14CeccarelliEclipse,AW15LeeYoung,AW1BinaryMapCastelli,AW2WorkSpaceSE3Jin,AW3MonteCarloWorkspacePeidro,AW4WorkspaceParallelAnnTanase,AW5WorkspaceParAnnGenKuzeci,AW6WorkspaceParaCampean,AW7WorkspaceParaAlp,AW8WorkspaceParWang,AW9RobotWorkspaceNumCao,AW10ParallelWrkspcOptHerrero2015,AW11SingMapParallelMacho,AW12HybridRobotModelPisla2013,AW13NonConvexWrkspcParallelHay2002}. 
Most of the current methods are focused on three-dimensional (3D) subspace which is projected from the original higher-dimensional manifold, because 3D is the maximum dimension that can be visualized \cite{AW2WorkSpaceSE3Jin}.


	
	
	
	
	
	


The focus of this paper is on improving the discretization method, one of the three major types of workspace algorithms. Discretization method involves dividing the desired spaces into partitions of equal volume, and further applying either forward or inverse kinematics. More specifically, 
the binary representation method, a variant of the family of discretization methods, involves the approximation of the workspace through the form of a binary map. The three-dimensional space is divided into multiple cuboid partitions with equal shape and volume, and an array is created to represent them. If a particular node is contained in the actual workspace, the value of the corresponding element will be set to 1, otherwise 0. A sequential initialization on all elements in the array generates the output.
 

However, this method time-consuming when the number of partitions is substantialy large. Take \cite{AW1BinaryMapCastelli}'s method for constant orientation workspace of all-revolute serial-link manipulators for example. Seeing the workspace algorithm as a mapping from a vector $\textit{\textbf{x}}$ that defines the robot and the scope to a bit vector $\textit{\textbf{y}} \in \mathbb{B}^n$ describing the workspace, the input vector is defined as the concatenation of the Denvait-Hartenberg parameters, coordinate values of all nodes on the partitions as well as the chosen orientation.


\begin{equation}
	\label{dh2}
	\textit{\textbf{y}}=f( \begin{bmatrix}
	\textit{\textbf{r}}_d & \textit{\textbf{r}}_a & \textit{\textbf{r}}_\alpha & \textit{\textbf{n}}_x & \textit{\textbf{n}}_y & \textit{\textbf{n}}_z & \textit{\textbf{n}}_i
	\end{bmatrix}^\top )
\end{equation}

The first part of the vector $\textit{\textbf{r}}_d , \textit{\textbf{r}}_a , \textit{\textbf{r}}_\alpha \in \mathbb{R}^d$ defines a manipulator regarding link offset, link length and link twist \cite{DHparameter}, where $d$ is the degree-of-freedom (DOF). $\textit{\textbf{n}}_x \in \mathbb{R}^{c_x+1}, \textit{\textbf{n}}_y \in \mathbb{R}^{c_y+1} , \textit{\textbf{n}}_z \in \mathbb{R}^{c_z+1} , \textit{\textbf{n}}_i \in \mathbb{S}^3$ describes the scope of the algorithm, where $c_x, c_y$ and $c_z$ are the number of elements in $\textit{\textbf{n}}_x, \textit{\textbf{n}}_y$ and $\textit{\textbf{n}}_z$ respectively. To promote memory efficiency, the technique of coordinate mapping is proposed: $\textit{\textbf{n}}_x, \textit{\textbf{n}}_y, \textit{\textbf{n}}_z$ respectively describe the Cartesian coordinate values an arbitrary node can take in the first, second and third dimension. To extract the specific coordinate values, one simply has to find all possible permutations of elements in $\textit{\textbf{n}}_x, \textit{\textbf{n}}_y, \textit{\textbf{n}}_z$, thus the name.  $\textit{\textbf{n}}_i$ describes the chosen orientation in terms of displacement around x, y and z axis, and the transformation from which to $SO(3)$ is also straightforward \cite{CorkeRobotics}: \par

\begin{equation}\label{rpy}
\textit{\textbf{R}}=\textit{\textbf{R}}_x(n_{i_1})\textit{\textbf{R}}_y(n_{i_2})\textit{\textbf{R}}_z(n_{i_3})
\end{equation}


In the implementation, the programmer could either choose to implement the forward kinematic equation or the inverse according to the specific situation. Due to the difficulty of the calculation of $\mathcal{K}^{-1}(\xi)$ and $\mathcal{K}(\textit{\textbf{q}})$, the algorithm has an expensive time complexity as $\textit{\textbf{x}}$ increases in length. For the inverse kinematics method, there are two factors that contributes to the time complexity. Either the robot itself becomes more complex (resulting in a longer vector describing it), or the number of points in the bit tensor is increasing (increasing scope or precision). Thus, the classical algorithm has a time complexity of

\begin{equation}\label{complexityBin}
	O((c(n))^3)+O(n^3)
\end{equation} 

, where $O(c(n))$ denotes the time complexity of the inverse kinematics algorithm as when the length of the vector describing the robot increases. The $O(n^3)$ term is resulted from the coordinate mapping method. Due to the abundance of matrix inversions, differential motion calculations and Jacobian computations in the inverse kinematic equation, the overall complexity is obviously worse than $O(n^3)$. Appendix \ref{complexity} shows a possible complexity of the algorithm when the method is implemented in MATLAB\textsuperscript \textregistered \space with depedent library created by \cite{CorkeRobotics}, and the time complexity for a general case is $O(n^{36})$. For the forward kinematics case, it is even harder to discuss the time complexity, since the process of sampling a joint configuration from probability distributions is unstable and time-consuming per se. 


Thus, it would be useful if an algorithm can compute the workspace without using either coordinates mapping or manipulator kinematic equations. Artificial neural networks, as universal approximators for high-dimensional non-convex functions \cite{MLPApproxHornik1989, MLPApproxHornik1990}, are found to be extremely helpful in this situation. In the following sections, the possibilities of developing an algorithm for manipulator workspace generation using techniques from deep learning are being explored. First, the problem of manipulator workspace will be formally defined. Then, a new technique termed subspace learning is introduced, exhibiting desirable characteristics from both classical methods and general deep learning algorithms. Finally, case studies on a class of 6-DOF manipulators is presented to demonstrate the feasibility of SL.

\begin{figure}[htb]
	\includegraphics[width=.5\textwidth]{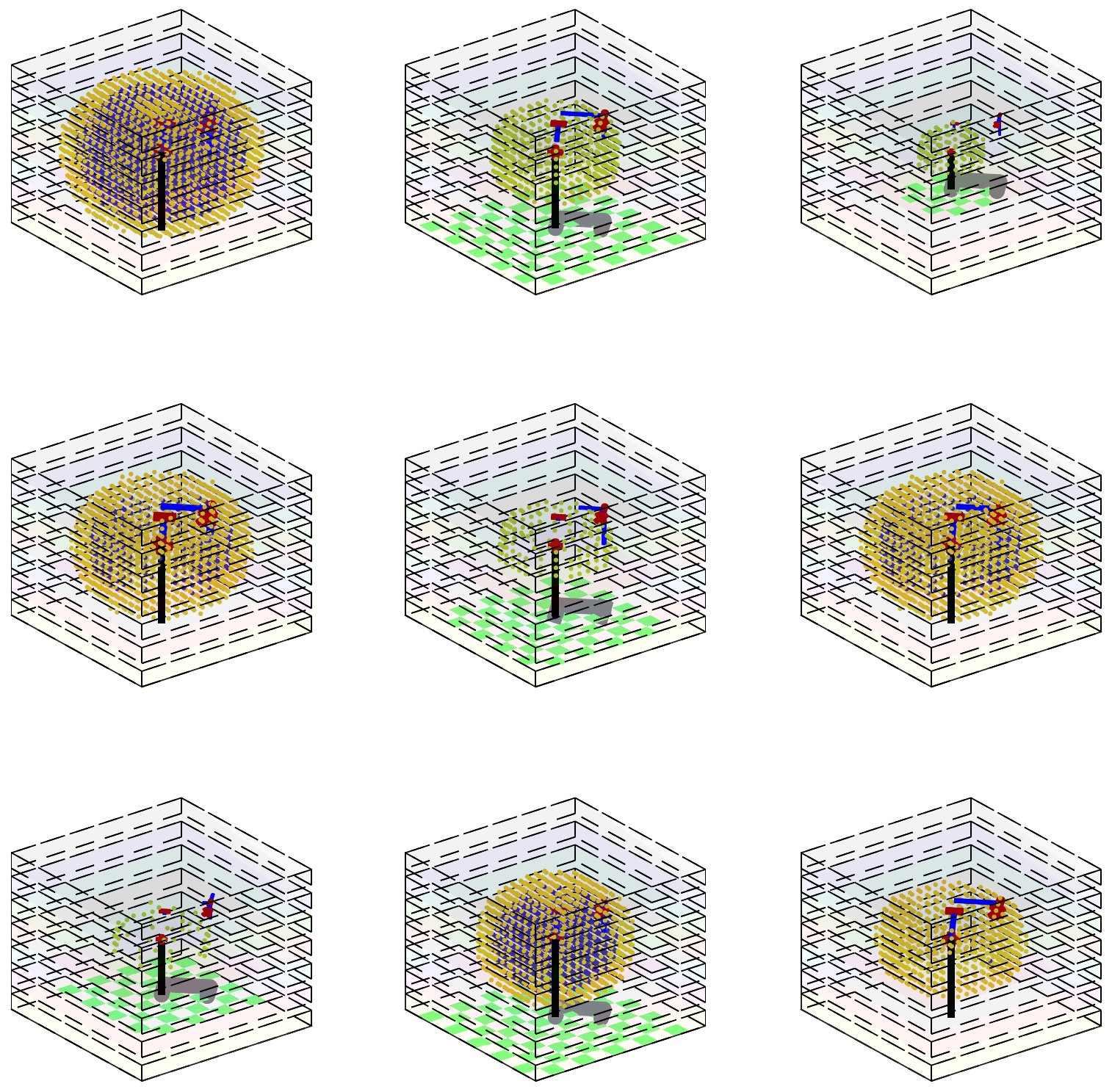}
	\caption{DNN-generated constant orientation workspace of 6-DOF manipulators with spherical wrist}
\end{figure}

\section{Related Works}
In the literature, the development of a neural network for workspace generation has been already tested on both serial-link and parallel mechanisms, utilizing several model architectures and optimization methods \cite{ KuzWorkspaceGeneticPar,FengAnnSerial}. The feasibility of using an artificial neural network for workspace analysis is first proposed and investigated by \cite{AW7WorkspaceParaAlp}, where a two-layer perceptron learns to generate the orientation workspace of a 6-3 SPM parallel mechanism based on a 3-tuple input. In a similar approach, \cite{AW6WorkspaceParaCampean} constructs a three-layer deep network that takes the input joint angles of a 2-DOF parallel manipulator and gives the workspace of the mechanism in a particular Cartesian plane. Levenberg-Marquardt method is used for optimization, and the lowest mean squared error obtained by the algorithm is 0.026.

\section{Problem Statement}
\subsection{Inverse Kinematics}
Since the purpose of this study is not to discuss different ways of obtaining the relationship between joint angle and end-effector velocities, and how it relates to existence of solutions (or lack thereof), only a general numerical solution is presented in Appendix \ref{ikine} \cite{NumericalInverseChiaverini}. For more information on this topic, one can refer to works of \cite{SicilianoRobotics} and \cite{CorkeRobotics}, where a wider spectrum of methods are discussed more in depth.

\subsection{Orientation and constant orientation workspace}
First, let $\textbf{\textit{q}}$ denote the joint configuration of a serial-link manipulator (for all-revolute manipulator $\textbf{\textit{q}} \in \mathbb{S}^n$) with a degree-of-freedom of $n$. $\textbf{\textit{a}} \in \mathbb{R}^3$ denotes the Cartesian coordinate of a point in 3-D space and $\textbf{\textit{R}} \in SO(3)$ denotes its orientation.
The pose of a manipulator (or any rigid body in general) encodes two different sets of information: position and orientation. Thus, the special Euclidean group $SE(3)$ is helpful in this context. It is clear in the form below that $SE(3)$ can store both types of information \cite{AW2WorkSpaceSE3Jin}:

\begin{equation}\label{se3}
SE(3) = \begin{Bmatrix}
(\textbf{\textit{a}},\textbf{\textit{R}})
\end{Bmatrix} = \mathbb{R}^3 \times SO(3)
\end{equation}

And the complete six-dimensional manipulator workspace:

\begin{equation}\label{workspace}
\mathbb{W} \subset SE(3) = \begin{Bmatrix}
(\textbf{\textit{a}},\textbf{\textit{R}})| \exists \textit{\textbf{q}}(\mathcal{K}^{-1}(\xi(\textbf{\textit{a}},\textbf{\textit{R}}))= \textit{\textbf{q}})
\end{Bmatrix}
\end{equation}
However, to store the workspace in a presentable manner, different variants of the complete workspace exist that put constraints either on position or orientation. 
 the orientation and constant orientation workspace of the manipulator, $\mathbb{W}_{ow} $ and $\mathbb{W}_{cow} $ respectively, can be defined as below:

\begin{equation}\label{orientationWorkspace}
\mathbb{W}_{ow} = \begin{Bmatrix}
\textbf{\textit{R}} \in SO(3) | \exists \textit{\textbf{q}}(\mathcal{K}^{-1}(\xi(\textbf{\textit{a}}_{const.},\textbf{\textit{R}}))= \textit{\textbf{q}})
\end{Bmatrix}
\end{equation}

\begin{equation}\label{constOrientWorkspace}
\mathbb{W}_{cow} = \begin{Bmatrix}
\textbf{\textit{a}} \in \mathbb{R}^3 | \exists \textit{\textbf{q}}(\mathcal{K}^{-1}(\xi(\textbf{\textit{a}},\textbf{\textit{R}}_{const.}))= \textit{\textbf{q}})
\end{Bmatrix}
\end{equation}

It is also possible to convert the workspace sets to tensors ($\fontfamily{phv}\selectfont \textbf{W}$,$\fontfamily{phv}\selectfont \textbf{W}_{ow}$, or $\fontfamily{phv}\selectfont \textbf{W}_{cow}$), for the sake of portability.
\subsection{Supervised Learning}
The problem of generating a workspace given manipulator and the scope parameter can be classified as a type of supervised learning, where  input predicts the "label". In machine learning, instead of using an analytic form of mapping from $\textit{\textbf{x}}$ to $\textit{\textbf{y}}$ for every value in a specific set (like $\mathbb{R}^n$), the rationale is probabilistic in essence. This means that knowledge about the input is presented as a probability distribution $p($\textbf{x}$)$, and the mapping is now the probability distribution of the targets \textbf{y} given \textbf{x} ( $p($\textbf{y} $|$ \textbf{x}$)$ ).


Thus, viewing the workspace as a vectorized random variable, we can arrive at a probabilistic model version of the workspace mapping:

Find $p_{ideal}($\textbf{y} $|$ \textbf{x}$;$ $\mathit{\bm{ \theta }})$, where $p($\textbf{x}$)$ describes the probability distribution of manipulator construction parameters and pose information of the scope. A state of $p_{ideal}($\textbf{y} $|$ \textbf{x}$;$ $\mathit{\bm{ \theta }})$ has elements describing whether a pose is within the true workspace and can be converted first to a bit tensor $\fontfamily{phv}\selectfont \textbf{P}$ and then to $\mathbb{W}$ defined above.

\section{Methodology}
\subsection{Redefining the problem}
The problem with the machine learning problem formulated above is that both \textbf{y} and \textbf{x} are infinite-dimensional. The probability distribution is designed to be capable of describing any manipulator with infinite precision, so the number of elements in the vectors will also be infinite. This is clearly unfeasible and unnecessary for real-world applications, so the first layer of simplification is to find instead a different distribution $p_{s}($\textbf{y}' $|$ \textbf{x}'$)$, that has random variables of finite dimensions. The new conditional distribution only presents the approximated workspace with finite precision given vectorized parameters of a fixed type of manipulators. To further make the problem suitable for deep learning, an altruistic method implemented in this paper is to treat the "scope" as concatenation of parameters regarding nodes on the finite partitions of either $\mathbb{R}^3$ or $SO(3)$. By taking advantage of Equation \eqref{dh2} \cite{AW1BinaryMapCastelli}, \textbf{x}' can be defined in the same way, except in terms of a random variable. $p_{s}($\textbf{y}' $|$ \textbf{x}'$)$ would then be a multivariate Bernoulli distribution describing whether what nodes are inside the real workspace. Thus, the discretization method mentioned in the introduction can now be seen as obtaining the state of \textbf{y}' given a state of \textbf{x}'. Direct assembly from a state of \textbf{x}' to the bit tensor $\fontfamily{phv}\selectfont \textbf{P}$ is impossible however, yet still feasible with informations from \textbf{y}' and specific code instructions.
This new distribution is in per se a different distribution compared to $p_{ideal}$, yet it is learnable by an artificial neural network, and it offers useful insights that enable us to extract information from the obtainable distribution.
\subsection{Experience Generation}
It is possible to obtain observations of the new objective distribution $p_{s}($\textbf{y}' $|$ \textbf{x}'$)$ by conventional workspace algorithm. The only modification is to disassemble the bit tensor $\fontfamily{phv}\selectfont \textbf{P}$ to a vector, so that the training of the dataset in the context of deep learning is possible. To design \textbf{x}', Equation \eqref{dh2} is used where altruistic precision, scope, and manipulator kinematic structure are chosen. In implementation, the process is online where values of specific elements are sampled from designed distributions. In this subspace learning scenario, which will be explained below, values are observations from different uniform distributions. The details of this can be found in Appendix \ref{experience}. 

 
\subsection{Subspace Learning}
Another problem about the application of deep learning in industrial scenarios is that artificial neural networks are merely approximators \cite{MLPApproxHornik1989, MLPApproxHornik1990}. In other words, it is impossible for deep learning algorithms to thoroughly learn the mapping \cite{NoFreeLunchWolpert}, and often challenging for them to reach high precision with a hard task and a limited computing capability.

A solution to the problem is subspace learning (SL), a deep learning algorithm based on subspace embedding that reduces the dimensionality of the distribution before the process of learning. The rationale behind SL is that sometimes not only are input-output pairs available to us, but also $p(\textbf{x})$ or even the true data generation process. Hence, learning the entire conditional distribution is often not necessary, and only some specific cases (or "subspaces," since sometimes they are in fact lower-dimensional subspaces of the original higher-dimensional space) of it are considered as the objective. So, instead of struggling to find the complete $p_{s}($\textbf{y}' $|$ \textbf{x}'$)$, it is simplified to

\begin{equation}\label{subspaceLearning}
	\begin{Bmatrix}
		\tilde{p}_{1}( \textbf{y}_{1}' | \textbf{x}_{1}' ) ,
		\tilde{p}_{2}( \textbf{y}_{2}' | \textbf{x}_{2}' ) ,
		\ldots ,
		\tilde{p}_{i}( \textbf{y}_{i}' | \textbf{x}_{i}' ) 
	\end{Bmatrix}
\end{equation}

, where $\tilde{p}_{1}( \textbf{y}_{1}' | \textbf{x}_{1}' ) , \tilde{p}_{2}( \textbf{y}_{2}' | \textbf{x}_{2}' ) , \ldots$are unnormalized modifications of the original $p_{s}($\textbf{y}' $|$ \textbf{x}'$)$ generated by altruistically altering $p(\textbf{x}_{1}'),p(\textbf{x}_{2}') \ldots$, $p(\textbf{x}_{1}',\textbf{y}_{1}'),p(\textbf{x}_{1}',\textbf{y}_{2}') \ldots$ or themselves. These modifications include but are not limited to: removing elements, 

\begin{equation}\label{removeElement}
\begin{split}
		 \textbf{x}_{i}' \subset \textbf{x}' \\
		 \textbf{y}_{i}' \subset \textbf{y}' 
\end{split}
\end{equation}

fixing elements,

\begin{equation}\label{fixElement}
\begin{split}
		\textbf{x}_{i}' = \begin{Bmatrix}
			$x$_{a}', \dots \sim  \delta(\mu_a) \dots | p_{s}($x$_{a}'=\mu_a)\neq 0, \dots \end{Bmatrix} \\ \cup \begin{Bmatrix} $x$ \in \textbf{x}' | $x$ \neq $x$_{a}, \dots
		\end{Bmatrix}
\end{split}
\end{equation}

 and changing parameters of a distribution, like the lower bound and the upper bound of a uniform distribution.
 
\begin{equation}\label{changeDistribution}
\begin{split}
	\textbf{x}_{i}' = \begin{Bmatrix}
			$x$_{a}', \dots \sim  \mathcal{U}(a_{min},a_{max}), \dots | a_{min},a_{max} \in \mathbb{R}, \dots \end{Bmatrix} \\ \cup \begin{Bmatrix} $x$ \in \textbf{x}' | $x$ \neq $x$_{a}, \dots
		\end{Bmatrix}
		\end{split}
\end{equation}

The last two modifications introduced here are not applicable to $\textbf{y}'$ since $p(\textbf{y}')$ is mostly unobtainable, while $p(\textbf{x}'|\textbf{y}')$ is the goal of learning. Thus, it is not logical to fix elements or change distribution type of some unknown distributions.

This reduces the dimensionality of the learning task even though the altered distributions are not used a priori. Although the changes are mostly done on $\textbf{x}'$, in some cases the structure of $\textbf{y}'$ might also be modified heuristically. For Equation \eqref{fixElement}, the new $\tilde{p}_{i}( \textbf{y}_{1}' | \textbf{x}_{1}' )$ is literally a geometrical subspace of the original high dimension distribution. While for examples like Equation \eqref{removeElement} and Equation \eqref{changeDistribution}, the new distributions only share elusive characteristics with its origin, and does not satisfy the literal meaning of "subspace."

Traditionally, only one approximator function $f_{original}(\textbf{\textit{x}};\bm{\mathit{\theta}})$ is created to learn and approximate the distribution, and sometimes the required capacity of $f_{original}$ is too large to be constructed and trained on existing platforms. Subspace learning, on the other hand, does not require the optimization of such function. A set of functions related to the subspaces are constructed instead that not only as a whole are analagous to the original function, but also have lower capacities.

\begin{equation}
\begin{split}
	& f_{original}(\textbf{\textit{x}};\bm{\mathit{\theta}}) \\
	& \begin{Bmatrix}
		f_{1}(\textbf{\textit{x}}_1;\bm{\mathit{\theta}}_1),f_{2}(\textbf{\textit{x}}_2;\bm{\mathit{\theta}}_2),...f_{i}(\textbf{\textit{x}}_i;\bm{\mathit{\theta}}_i)
	\end{Bmatrix}
\end{split}
\end{equation}

This is where the trade-off comes in. Intuitively, the total time of optimizing $f_{original}$ is much less than optimizing all possible subspace functions $f_1,f_2 \ldots f_n$, if the means to create a subspace is properly defined. However, for certain tasks the majority of values in $\textbf{\textit{x}}$ is never used to calculate $f_{original}(\textbf{\textit{x}};\bm{\mathit{\theta}})$, but whatever are being used does belong to the same distribution. Thus, the subspace learning alternative can be used to fix the values of those elements.
Moreover, if a specific machine learning task only requires a small number of subspaces, the sequential training of subspaces is more feasible and require less time than training 
the original function.



\begin{figure}[!htb]
	\centering
	\includegraphics[width=.5\textwidth]{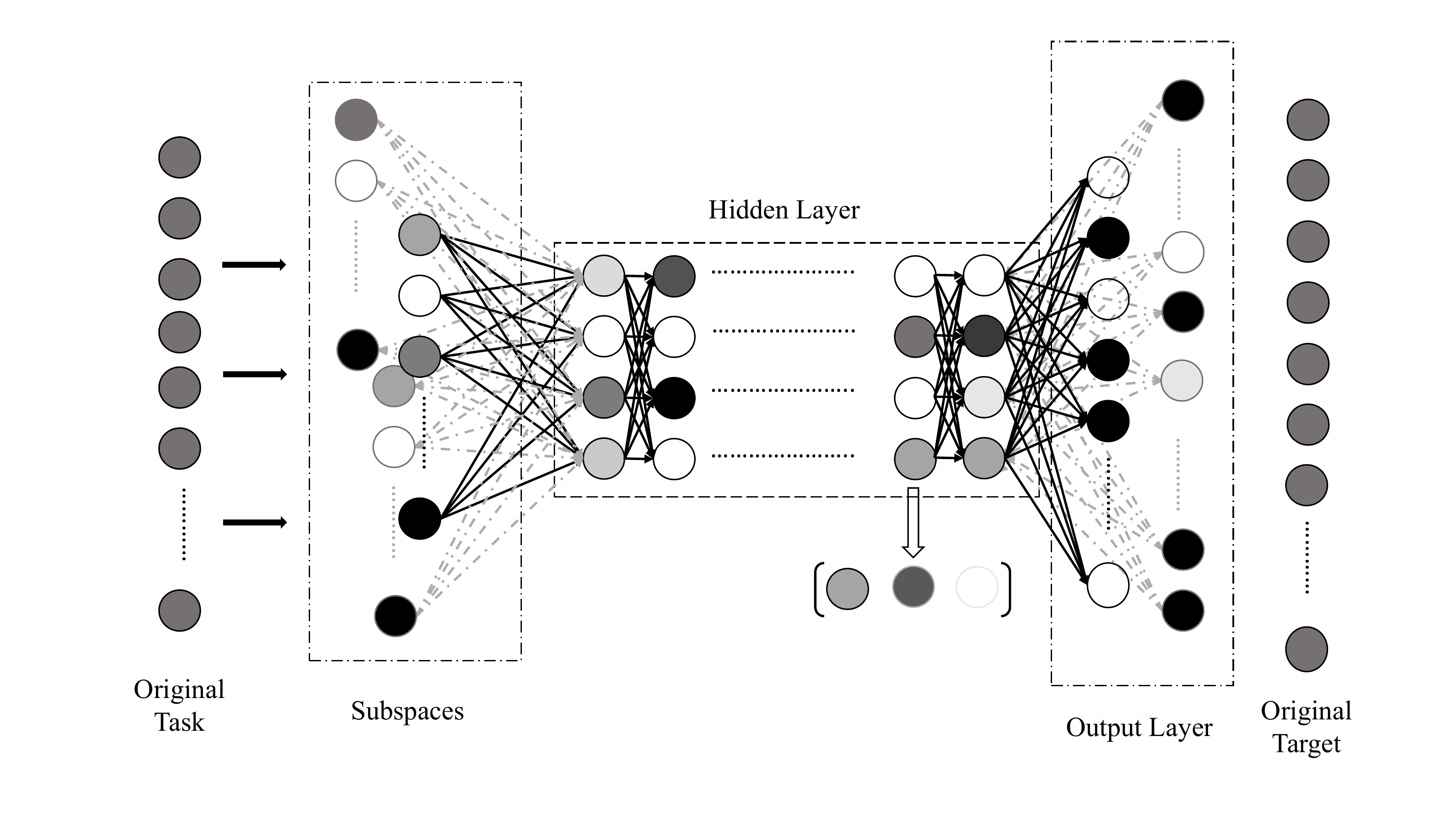}
	\caption{Network Architecture}
	\label{fig:mesh2}
\end{figure}

Figure \eqref{fig:mesh2} presents the DNN model of subspace learning. There are two remarks on the architecture. First, the original samples are never used for the actual training of the algorithm. Instead, the optimization process is done on various extracted subspaces. The second remark is that all subspace models use the same hidden layer architecture. Apparently, the values of parameters are different if the same model is optimized to suit specific subspaces. To address the problem, all neurons will store multiple sets of parameters. During the learning process, different subspaces are optimized by different altruistic methods and then stored in the same model. During run-time, the system will go through a linear-search to make the neurons take the specific set of parameters tuned especially for the input. 




\subsubsection{Distinctions between Subspace Learning and Generic Feedforward Networks}
There are a few nuances between the DNN model for subspace learning and a typical deep feedforward network. First, there are multiple goals for optimization in subspace learning, although optimizing the parameters of a specific subspace is the same as that of a feedforward network. The cost function is now the average loss over the subspace instead of the original training set, which can be written as (for the $i$-th subspace)

\begin{equation}\label{loss}
	J(\mathit{\bm{ \theta }}_i)= \mathbb{E}_{(\textit{\textbf{x}},\textit{\textbf{y}}) \sim p_i}L(f_i(\textit{\textbf{x}};\mathit{\bm{ \theta }}_i),\textit{\textbf{y}})
\end{equation}

And, applying the concept of empirical risk minimization with $m$ examples \cite{GoodfellowDeep}, 

\begin{equation}\label{empiricalRiskSubspace}
\begin{split}
& \mathbb{E}_{({\textbf{\textit{x}}}_i,{\textbf{\textit{y}}}_i) \sim \hat{p}_i}L(f_i(\textbf{\textit{x}}_i;\mathit{\bm{ \theta }}_i),\textbf{\textit{y}}_i) \\
= & \frac{1}{m}\sum_{i=1}^{m} [L(f_i(\textit{\textbf{x}}^{(i)}_i;\mathit{\bm{ \theta }}_i),\textit{\textbf{y}}^{(i)}_i)]
\end{split}
\end{equation}

\begin{equation}\label{empiricalRisk}
\begin{split}
& \mathbb{E}_{(\textit{\textbf{x}},\textit{\textbf{y}}) \sim p}L(f_{original}(\textbf{\textit{x}};\mathit{\bm{ \theta }}),\textbf{\textit{y}}) \\  =  & \mathbb{E}_{(\textit{\textbf{\textit{x}}}_n,\textit{\textbf{\textit{y}}}_n) \sim p_n} [\mathbb{E}_{\textbf{m} \sim p_{subspace}} [L(f_n(\textbf{\textit{x}}_n;\mathit{\bm{ \theta }}_n),\textbf{\textit{y}}_n)]] \\
 \approx & \mathbb{E}_{(\textit{\textbf{\textit{x}}}_n,\textit{\textbf{\textit{y}}}_n) \sim \hat{p}_n} [\mathbb{E}_{\textbf{m} \sim p_{subspace}} [L(f_n(\textbf{\textit{x}}_n;\mathit{\bm{ \theta }}_n),\textbf{\textit{y}}_n)]] \\
\end{split}
\end{equation}


Equation \eqref{empiricalRisk} presents a loosely defined expectation of loss expectations of all possible subspaces taken over first the random variable parameters defining the subspace $\textbf{m}$ then the model distribution (and thus the empirical distribution). It is believed that if the ways to create subspace is limited to a few well-defined procedures, the expectation is equal to the original expectation of the loss of the would-be created function. This is, however, not proven mathematically, and hence more theoretical works beyond the example given in Appendix \ref{proof} are required. The only certainty is that all subspaces provide useful characteristics that are analogous to that of the original one. 
Notice that the expectation is taken from the subspace empirical distribution instead of the entire data generating empirical distribution $\hat{p}_{s}$. 

For regularized models of a subspace, 

\begin{equation}\label{regularization}
	J_{i}'(\mathit{\bm{ \theta }}_i; \mathbb{X}_i,\mathbb{Y}_i)= J_{i}(\mathit{\bm{ \theta }}_i; \mathbb{X}_i,\mathbb{Y}_i)+\beta_i \Omega_i(\mathit{\bm{ \theta }}_i)
\end{equation}

The regularization function $\Omega(\mathit{\bm{ \theta }})$ and the hyperparameter $\beta \in [0,+\infty)$ can be the same for every subspace, or it can be altered altruistically to adapt to specific situations. 

Intuitively, all existing optimization and regularization methods are applicable to train the subspace, and in the case studies section results between some of the most popular optimization methods are compared to show some of the nuances in subspace optimization.

The second major distinction is that training a SL algorithm includes the integration of different models. Once traditional optimization methods are done for every subspace and optimal parameters are obtained, it is possible to construct a single model to store all the information. Traditionally, two graphs will be created for a feed-forward neural network, one graph $\mathcal{G}$ for forward propagation and a subgraph of the former $\mathcal{B}$ for back-propagation \cite{GoodfellowDeep}. In subspace learning, two computational graphs are also created for a subspace, except for the following differences. First, instead of storing weight matrices for the hidden layer, tensors are created where $\fontfamily{phv}\selectfont \textbf{W}_{:,:,i}^{(j)}$, $j \in \begin{Bmatrix} 1,2,....l \end{Bmatrix}$ denotes the $i$-th subspace weight matrices of a model with a depth of $l$. The same is true for biases, where

\begin{equation}\label{biasMatrix}
    \textbf{\textit{B}}_{:,i}^{(j)} , j \in \begin{Bmatrix} 1,2,.....l \end{Bmatrix}
\end{equation}

denotes the bias parameters of the $i$-th subspace in the same model. The information of input/output parameters are stored in other altruistic data structure. The second difference has to do with the connections between the hidden layer and the input/output layers. When a specific subspace is used or being trained, a new graph $\mathcal{G}_{new}$ will be created at runtime that incorporates the input/output layer and the hidden layer. Since different subspaces may have different input and output vector length, $Pa_{\mathcal{G}_{new}}(\textbf{\textit{u}}^{(2)})$ differs from subspaces to subspaces. The same is also true for the last layer $\textbf{\textit{u}}^{(l)}$. Except for the first and last layers, the new graph's layers contain exactly one edge for an edge connecting two arbitrary nodes in $\mathcal{G}$. Rest of the connections are made exactly like a generic fully-connected feed-forward network.

\subsubsection{Context in Manipulator Workspace problem}
\begin{figure}[!htb]
	\centering
	\includegraphics[width=.5\textwidth]{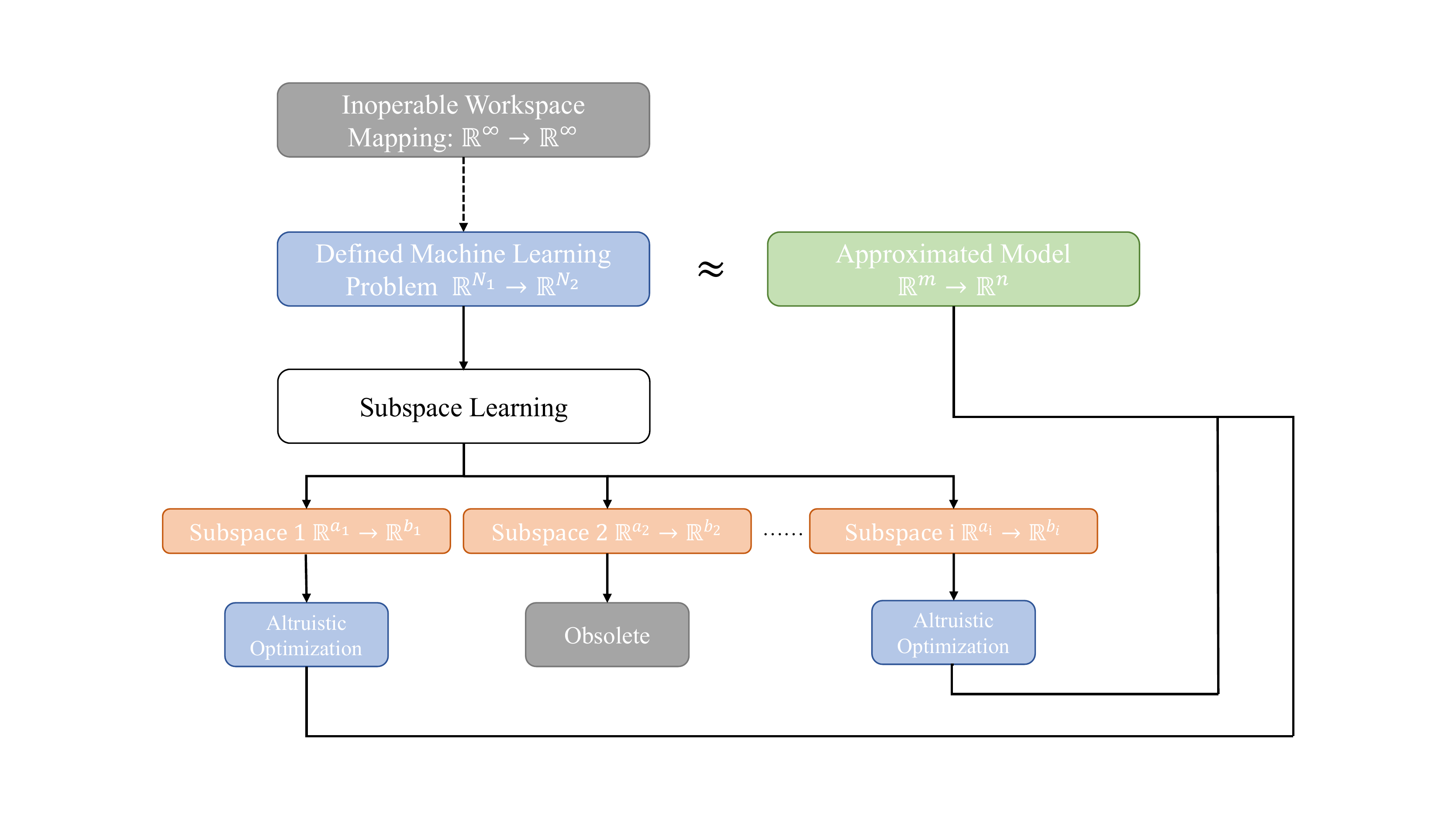}
	\caption{Structure of subspace learning in manipulator workspace problem}
	\label{fig:mesh4}
\end{figure}

The motivation of creating a subspace learning network is mainly to addresses needs in the field of robotics, so it is especially useful in manipulator workspace calculation for the following. First, although Denvait-Hartenberg parameters are capable of describing a broad spectrum of kinematic structures, only a few have real-world applications. For example, spherical wrists significantly enhance the manipulability of robots because it allows rotation of the end-effector around a single point in three-dimensional space. 
Now consider an element in the input vector $x_i$ that represents a link twist between two links of a manipulator. Instead of learning the complete distribution x $\sim \mathcal{U}_{\mathbb{S}}$, which is unnecessary because the algorithm has to adjust for peculiar angles like $\frac{2\pi}{13}$, three subspaces are created to the algorithm where for a subspace the algorithm observes a fixed state of x (like x=$\frac{\pi}{2}$, $0$ or $-\frac{\pi}{2}$). Graphically, the algorithm is learning a hyperplane in the original higher-dimensional space. This subspace will fail miserably if the $x_i$ in the input does not equal to any of the recorded values since optimization and regularization measures only tailored the algorithm given chosen inputs. However, because most of the input in real-world applications will be one of the recorded values, this induced-overfit method makes the algorithm easier to train with higher precision.
Figure \eqref{fig:mesh4} presents the structure of subspace learning algorithm when applied to workspace calculation.

\subsubsection{Time Complexity}
The time complexity of subspace learning is defined as the asymptotic behavior of the runtime of the forward prorogation graph as the length of the vector describing a subspace increases. It does not concern with the training time. Since the connections within the hidden layer are constant, assuming all operations between two nodes take the same time, a linear increase in the length of input vector only causes a linear increase in the connections between the first two and a cubic increase for the last two layers. More specifically, if $k$ connections already exist in $\mathcal{G}_{new}$ and $a_1$ new elements are added to the input vector that leads to $a_2$ new elements added to the output layer. If there are $m_1$ nodes in the first layer and $m_2$ nodes in the last layer of the graph, the number of connections after the increase can be expressed as:

\begin{equation}\label{knew}
    k_{new}=k+a_{1}m_{1}+a_{2}m_{2}
\end{equation}   

Assuming constant time of the computations between nodes in the graph, the run-time of the algorithm can be also expressed as a function of the input vector length $n$:

\begin{equation}\label{fn}
	f(n)=c_{core}+c_{1}n+c_{2}g(n)
\end{equation}

Where $g(n)$ is a polynomial in $n$ with a degree of 3. This exists because subspace learning still has characteristic analogous to coordinate mapping used in classical methods.


Over the long run, this has a time complexity of $O(n^3)$. Although on the first sight it's still expensive, locally subspace learning drastically decreases the run time by substituting complex operations by the simplest ones like addition, multiplication and exponentiation.

\section{Experimental Setup: 6-DOF Serial-link Manipulator with Spherical Wrist}
The 6-DOF serial-link manipulator with a spherical wrist is arguably the most prevalent type of serial-link manipulators in the industry due to its high flexibility and the existence of an analytic solution to the inverse kinematics. The kinematic structure of this type of robot is included below:
\begin{center}
\begin{tabular}{|c|c|c|c|c|}
	\hline
	$j$ & $\theta$ & $d$ & $a$ & $\alpha$ \\
	1 & $q_1$ & 0 & 0 & $\frac{\pi}{2}$ \\
	2 & $q_2$ & 0 & $a_1$ & 0 \\
	3 & $q_3$ & $d_1$ & $a_2$ & $-\frac{\pi}{2}$ \\
	4 & $q_4$ & $d_2$ & 0 & $\frac{\pi}{2}$ \\
	5 & $q_5$ & 0 & 0 & $-\frac{\pi}{2}$ \\
	6 & $q_6$ & 0 & 0 & 0 \\
	\hline
\end{tabular}
\end{center}


In the table, $d_1$,$d_2$ and $a_1$ describes the length of three major parts of the manipulator, and $a_2$ describes the shoulder offset that may not apply to some of the variants.
To transfer the manipulator worksapce problem to a machine learning problem, the design of the input distribution $p(\textbf{x})$, and its corresponding subspaces are important. The remark here is that there are two separate steps in this process: defining a solvable machine learning problem from the inoperable infinite workspace mapping, and creating its subspaces. The former includes setting the general scope (coordinate information of the nodes on the cuboid partitions and precision) and the degree-of-freedom of the manipulator (the length of $\begin{bmatrix}\textit{\textbf{r}}_d & \textit{\textbf{r}}_a & \textit{\textbf{r}}_\alpha \end{bmatrix}$), while the latter specifies the exact kinematic structure and sometimes crop part of the original output for the sake of dimensionality reduction in training.

For the specification of the kinematic model:

\begin{equation}
\begin{split}
\textbf{x} & = \begin{bmatrix}
	\textbf{r}_d & \textbf{r}_a & \textbf{r}_\alpha & \textbf{n}_x & \textbf{n}_y & \textbf{n}_z & \textbf{n}_i
	\end{bmatrix}^\top \\
	r_{d_i} &\sim \mathcal{U}(0,\beta) |i \in \begin{Bmatrix} 3,4 \end{Bmatrix} \\
	r_{d_i} &\sim \delta(0) |i \in \begin{Bmatrix} 1,2,5,6\end{Bmatrix} \\
	r_{a_i} &\sim \mathcal{U}(0,\beta) |i \in \begin{Bmatrix} 2,3 \end{Bmatrix} \\
	r_{a_i} &\sim \delta(0) |i \in \begin{Bmatrix} 1,4,5,6\end{Bmatrix} \\
	r_{\alpha} &\sim \begin{bmatrix}  \delta(\frac{\pi}{2})& \delta(0)& \delta(-\frac{\pi}{2})& \delta(\frac{\pi}{2})& \delta(-\frac{\pi}{2})& \delta(0) \end{bmatrix}^\top 
\end{split}
\end{equation}

For constant orientation workspace,

\begin{equation}
\begin{split}
	n_{x_k} &\sim \delta(\textit{n}_{x_k})\\
	n_{y_k} &\sim \delta(\textit{n}_{y_k})\\
	n_{z_k} &\sim \delta(\textit{n}_{z_k})\\
	n_{i_k} &\sim \delta(\textit{n}_{i_k})\\
	\textbf{n}_x &\in \mathbb{R}^{l_x} | -2\beta \leq n_{x_i} \leq 2\beta , n_{x_{i+1}}-n_{x_{i}} = \Delta x \\
	\textbf{n}_y &\in \mathbb{R}^{l_y} | -2\beta \leq n_{y_i} \leq 2\beta , n_{y_{i+1}}-n_{y_{i}} = \Delta y \\
	\textbf{n}_z &\in \mathbb{R}^{l_z} | -2\beta \leq n_{z_i} \leq 2\beta , n_{z_{i+1}}-n_{z_{i}} = \Delta z \\
	\textbf{n}_i &\in \mathbb{S}^3 \\
\end{split}
\end{equation}

All permutation of states in $\textbf{n}_x$, $\textbf{n}_y$ and $\textbf{n}_z$ defines a point in $\mathbb{R}^3$, and the states in $\textbf{n}_i$ defines a point in $SO(3)$ according to Equation \eqref{rpy}.

$\beta$ is an arbitrary value in $\mathbb{R}$. It does not effect the algorithm since scaling the D-H parameter of a manipulator will result in the scaling of the $\mathbb{R}^3$ part of the workspace in the same manner. In the training sessions of this paper, $\beta$ is set to be $0.5$.

For orientation workspace, the input vector is defined in a similar way except for a few differences shown below:

\begin{equation}
\begin{split}
	\textbf{n}_x &\in \mathbb{S}^{l_x} | -2\beta \leq n_{x_i} \leq 2\beta , n_{x_{i+1}}-n_{x_{i}} = \Delta x \\
	\textbf{n}_y &\in \mathbb{S}^{l_y} | -2\beta \leq n_{y_i} \leq 2\beta , n_{y_{i+1}}-n_{y_{i}} = \Delta y \\
	\textbf{n}_z &\in \mathbb{S}^{l_z} | -2\beta \leq n_{z_i} \leq 2\beta , n_{z_{i+1}}-n_{z_{i}} = \Delta z \\
	\textbf{n}_i &\in \mathbb{R}^3 \\
\end{split}
\end{equation}

The design of the input vector in the spherical wrist case can be seen as a subspace of the machine learning problem defined in a generic 6-DOF manipulator, where states of $\textbf{r}_d , \textbf{r}_a $ are in $\mathbb{R}$ and states of $\textbf{r}_\alpha$ are in $\mathbb{S}$. 


Deep neural networks are constructed in MATLAB\textsuperscript \textregistered \space R2017b with functions of feedforward networks in the Neural Network Toolbox. Experience are generated in the same environment as well with the implementation of algorithm in Appendix C. Results are compared between different subspaces of the two types of workspaces, different depth and width, different levels of stochasticity, and different methods of optimization and regularization. Since the outputs of the network are in $\mathbb{R}^n$ while the ideal states are in $\mathbb{B}^n$, a simple fine-tuning function is created to cluster the outputs into $0$s and $1$s. A part of the discussion is focused on the choice of the threshold value in this function. Also, a typical subspace learning algorithm is compared to the classical method in terms of run-time.

\section{Discussion}
\subsection{Analysis of a particular Subspace}

\begin{figure}[htb]
	\centering
	\includegraphics[width=.5\textwidth]{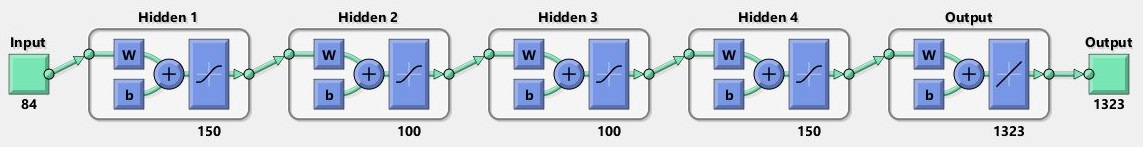}
	\caption{Feedforward neural network}
	\label{fig:mesh10}
\end{figure}

\begin{figure}[h]
	\centering
	\includegraphics[width=0.5\textwidth]{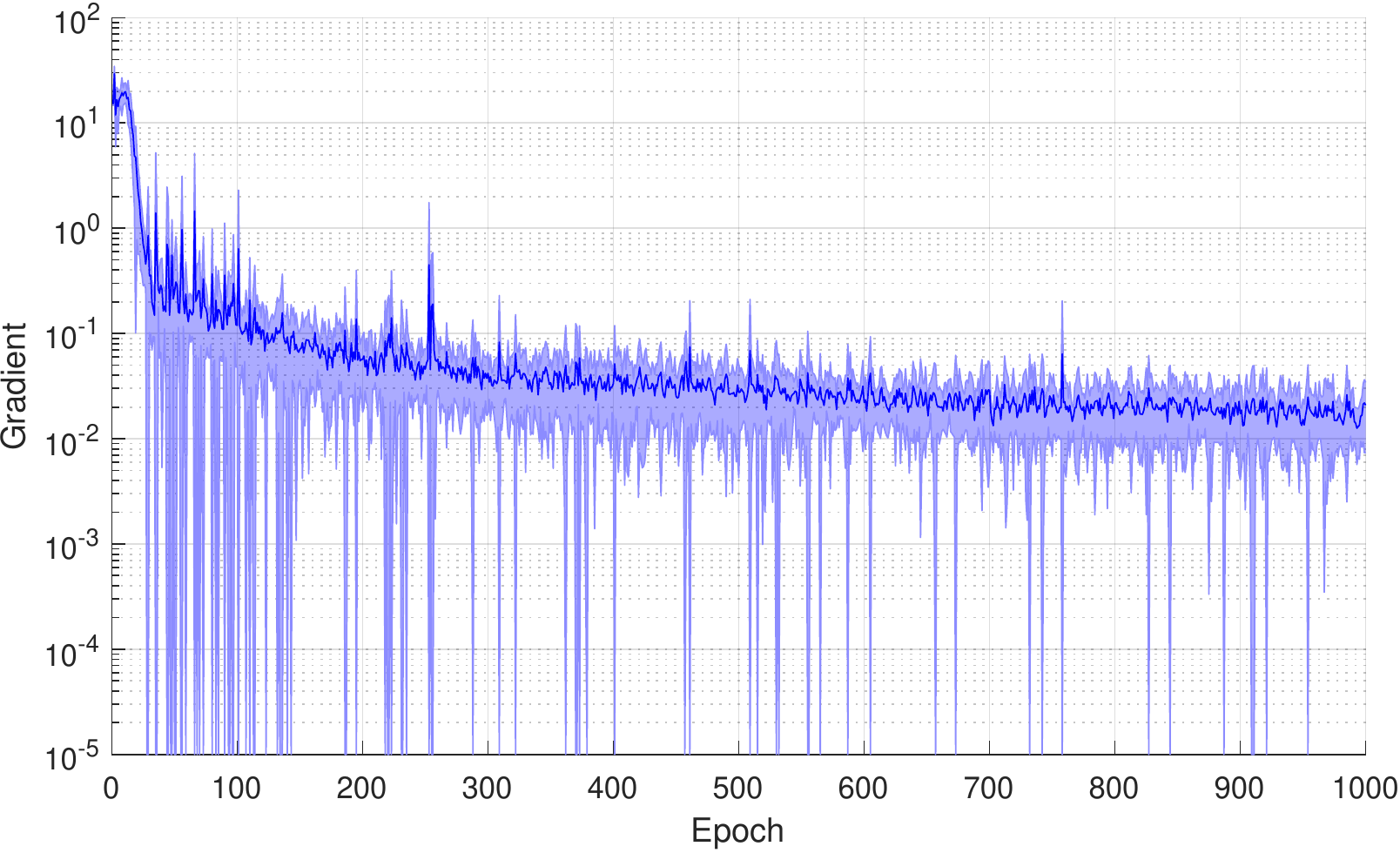}
	\caption{Average gradient ($\pm \sigma$) of 10 training sessions}
	\label{fig:mesh6}
\end{figure}

The first subspace is constructed to approximate the constant orientation workspace mapping of a class of 6-DOF serial-link spherical wrist manipulators. The x,y,z-range are all on the interval $[-1,1]$, with $\Delta x = \Delta y =\Delta z =0.1$. The orientation for the constant orientation workspace is $[0, 0,0]$, which translates to $\textbf{I}_3$ in terms of $SO(3)$. The embedding takes place both in the feature space and the output space, where for the latter only $5293$-th to $6615$-th elements are selected to be learnt by the deep neural network. This represents the middle section of the original cuboid partitions. A four-layer fully connected feedforward network is constructed in MATLAB\textsuperscript \textregistered , with hyperbolic tangent function as activation function and an additional affine layer before output. Figure \eqref{fig:mesh10} shows the architecture of the neural network. Training of the parameters is done by a batch descent of $1 \times 10^5$ samples, and the implementation of \cite{RpropRiedmiller}'s resilient backpropogation. The batch is then divided randomly into training set, validation set and testing set, with the ratio of $0.7:0.15:0.15$. Figure \eqref{fig:mesh6} presents the change in gradient during the first 200 epochs.

\begin{figure}[h]
	\centering
	\includegraphics[width=.5\textwidth]{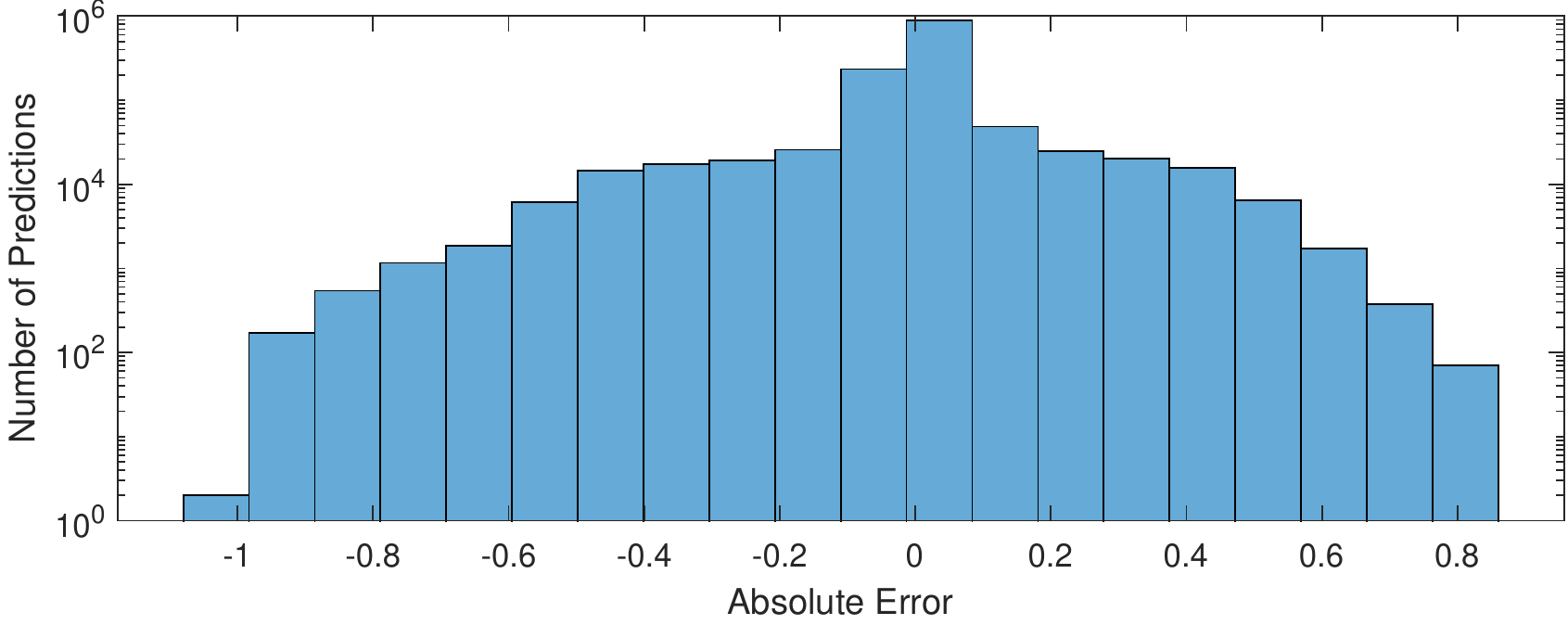}
	\caption{Average error of predictions on a test set of 1000 examples}
	\label{fig:mesh7}
\end{figure}

\begin{figure}[h]
	\centering
	\includegraphics[width=.5\textwidth]{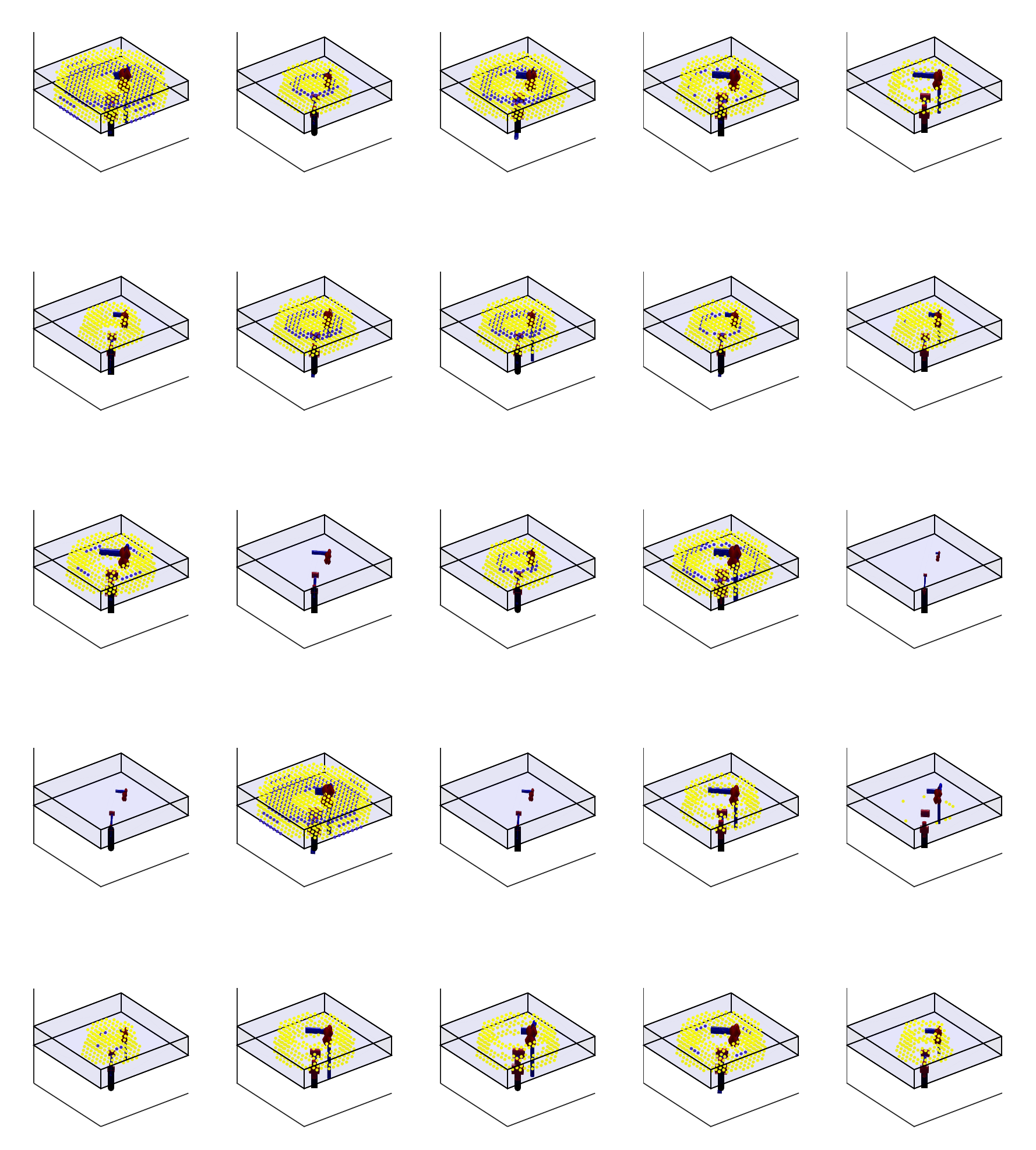}
	\caption{DNN-generated workspace section of 25 random manipulators in the test set}
	\label{manipulators}
\end{figure}

The subspace embedding reaches high accuracy during the first 200 gradient updates and continues to converge. The result is repeated for 10 training sessions, each with $10^3$ epochs done on the training set. Figure \eqref{fig:mesh7} presents the prediction error of an independent test set of $10^3$ samples evaluated on the best performing session after training. With a filter function of threshold value $0.5$, the network is able to predict over $99\%$ of the samples correctly. On average, the network has a F-measure of $0.9665$ with a standard deviation of $0.0045$.

 

\subsection{Different subspaces with the same dimension}
\begin{figure}[!h]
	\centering
	\includegraphics[width=.5\textwidth]{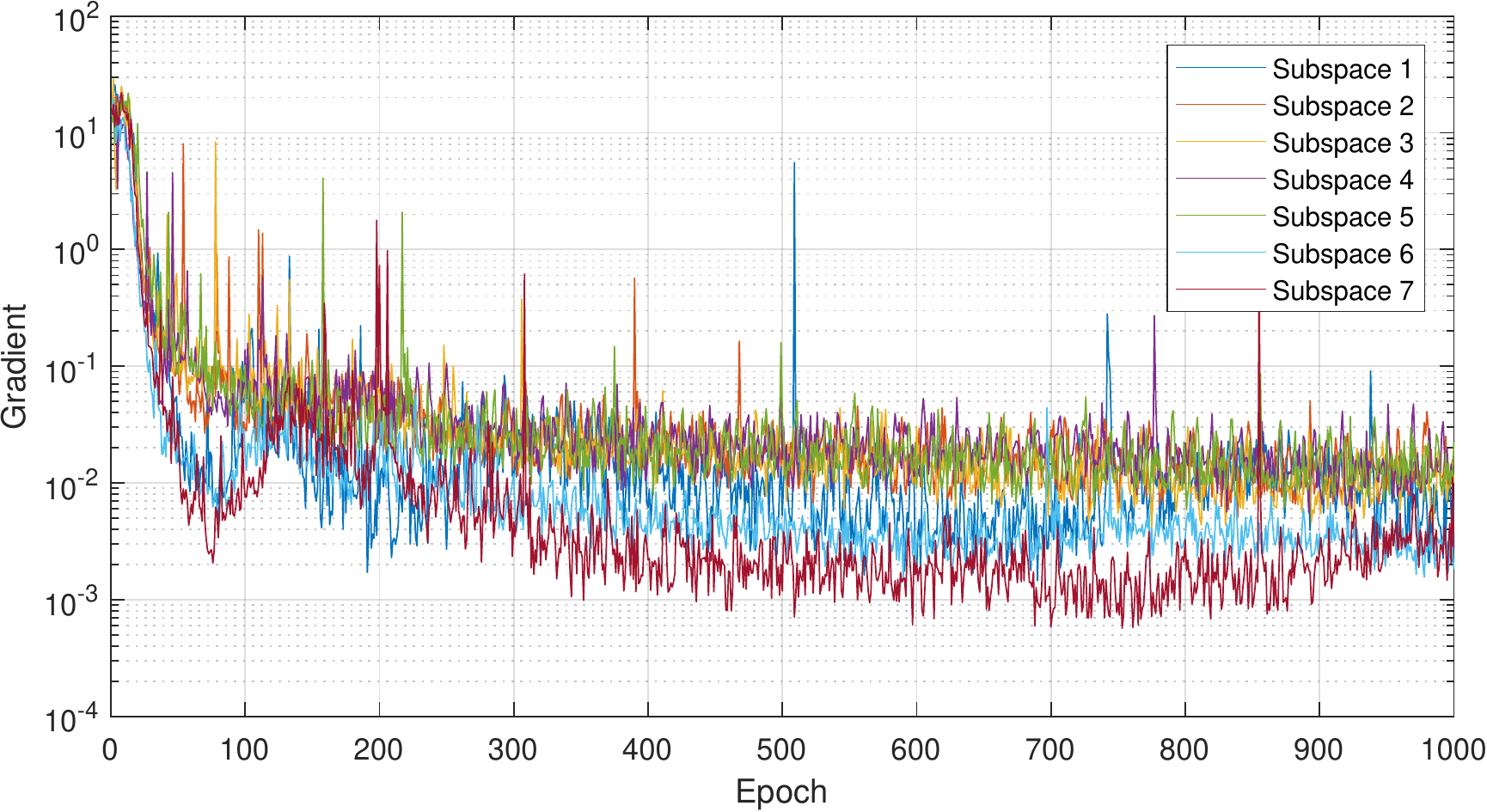}
	\caption{Subspace training of equal dimension}
	\label{fig:mesh11}
\end{figure}

\begin{figure}[!h]
	\centering
	\includegraphics[width=.5\textwidth]{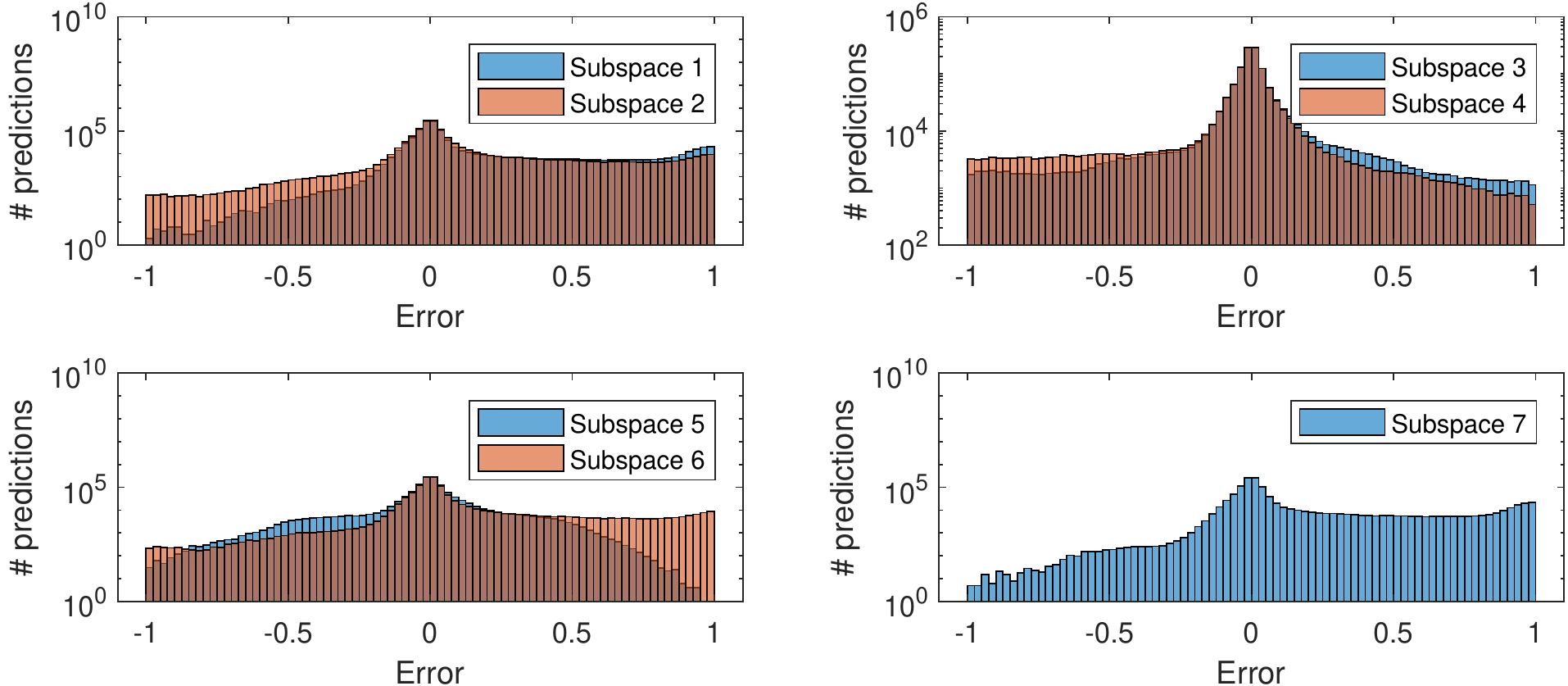}
	\caption{Histogram of prediction errors}
	\label{fig:mesh12}
\end{figure}

\begin{figure}[!h]
	\centering
	\includegraphics[width=.5\textwidth]{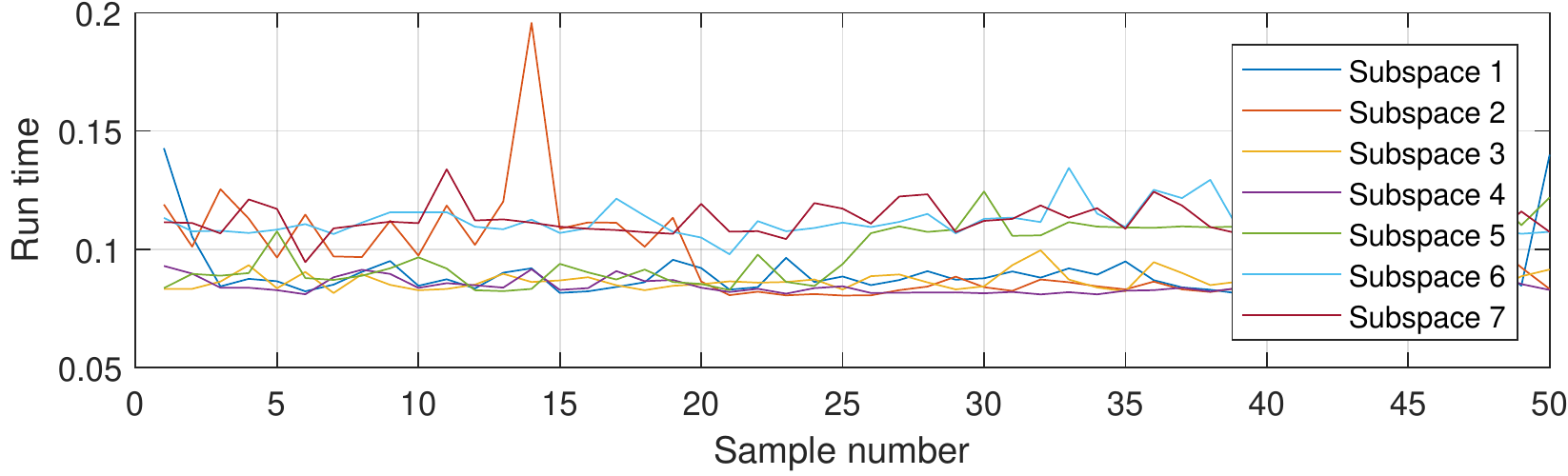}
	\caption{Run time of subspace models}
	\label{fig:mesh13}
\end{figure}

To verify the presumption that subspaces with the same size are similarly easy to train, seven subspace models are created with equal dimensionality according to section 6.1, except that each subspace specialize in the mapping of a particular portion in $\mathbb{R}^3$. The network still has a depth of $4$, except that the number of neurons in the first and last hidden layer is reduced to $125$. Figure \eqref{fig:mesh11} shows that some subspaces suits better to the optimization method than others when trained on a batch of $6 \times 10^4$ samples, which highlights the need of embedding to apply subspace-specific methods. Furthermore, Figure \eqref{fig:mesh12} suggests that with the same gradient updates, the middle subspaces has a higher precision, possibly due to the sparse nature of positives in nodes far away from the origin of the manipulator. Figure \eqref{fig:mesh13} then confirms that the computer run time of trained network does not deviate significantly between subspaces.

\begin{figure}[!h]
	\centering
	\includegraphics[width=.5\textwidth]{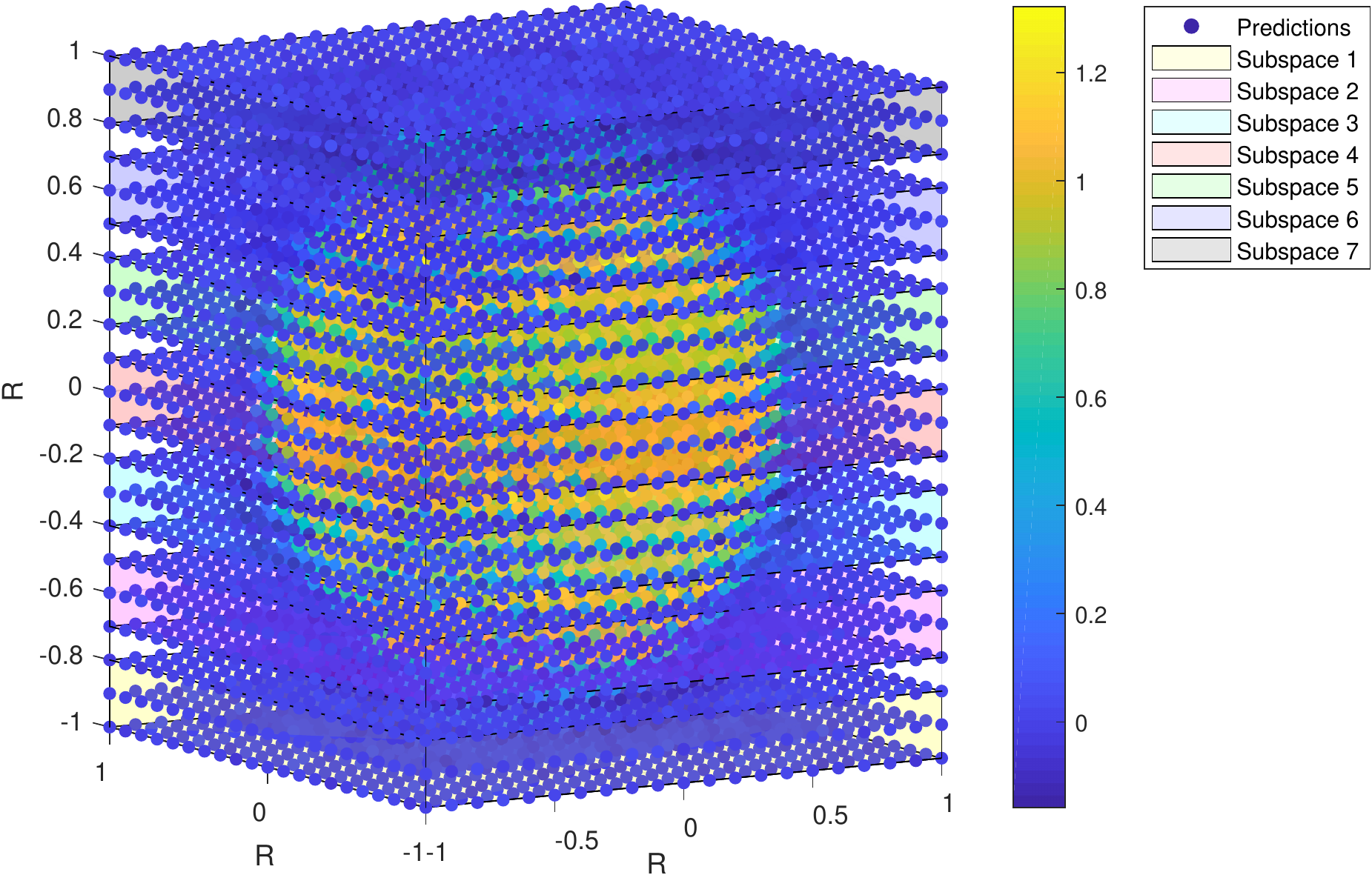}
	\caption{Workspace of PUMA560 robot generated by 7 subspace models}
	\label{fig:mesh14}
\end{figure}

Subspace learning enables models with small hypothesis space to mimic higher-dimensional distributions with several sets of parameters. In this case, it is possible to generate the full cuboid constant orientation workspace of the manipulator from the collaboration of the 7 subspaces that were trained separately. Figure \eqref{fig:mesh14} presents the deep-neural-network-generated result of PUMA560 workspace.

\subsection{Other Workspaces}
\begin{table}[!h]
\caption{F-measures of other subsapce models}
\label{f-workspace}


\begin{center}
\begin{tabular}{cccccc}
\multicolumn{1}{c}{\bf Workspace} &\multicolumn{1}{c}{\bf $\Delta$} & {\bf Size of $y \in \mathbb{B}^n$} & \multicolumn{1}{c}{\bf F-measure ($\pm \sigma_f$)} & {\bf Batch Size} 
\\ \hline \\
 $\mathbb{R}^3$ & $1$ & $27$& $1.000 \pm 0.00$ & $6 \times 10^4$\\
 $\mathbb{R}^3$ & $0.5$ & $125$& $0.9967 \pm 0.03$ & $6 \times 10^4$\\
 $\mathbb{R}^3$ & $0.1$ & $1323$& $0.9221 \pm 0.14$ & $6 \times 10^4$\\
 $\mathbb{R}^3$ & $0.1$ & $3086$& $0.8493 \pm 0.19$ & $3 \times 10^4$\\
 $\mathbb{R}^3$ & $0.1$ & $9261$& $0.9038 \pm 0.15$ & $1.5 \times 10^4$\\
 $SO(3)$ & $\frac{\pi}{3}$ & $343$& $0.9981 \pm 0.01$ & $6 \times 10^4$\\
 $SO(3)$ & $\frac{\pi}{8}$ & $4913$& $0.9779 \pm 0.13$ & $1.5 \times 10^4$\\
 $SO(3)$ & $\frac{\pi}{10}$ & $9261$& $0.9920 \pm 0.03$ & $1.5 \times 10^4$\\
\end{tabular}
\end{center}
\end{table}

According to Table 1, for both constant orientation workspace and orientation workspace, the model reaches relatively high precision in predicting the outputs. Though beyond the scope of this paper, it is believed that a deep neural network is also capable of predicting the constant orientation workspace of an underactuated serial-link manipulator.

\subsection{Computer Run-time}
\begin{figure}[!h]
	\centering
	\includegraphics[width=.5\textwidth]{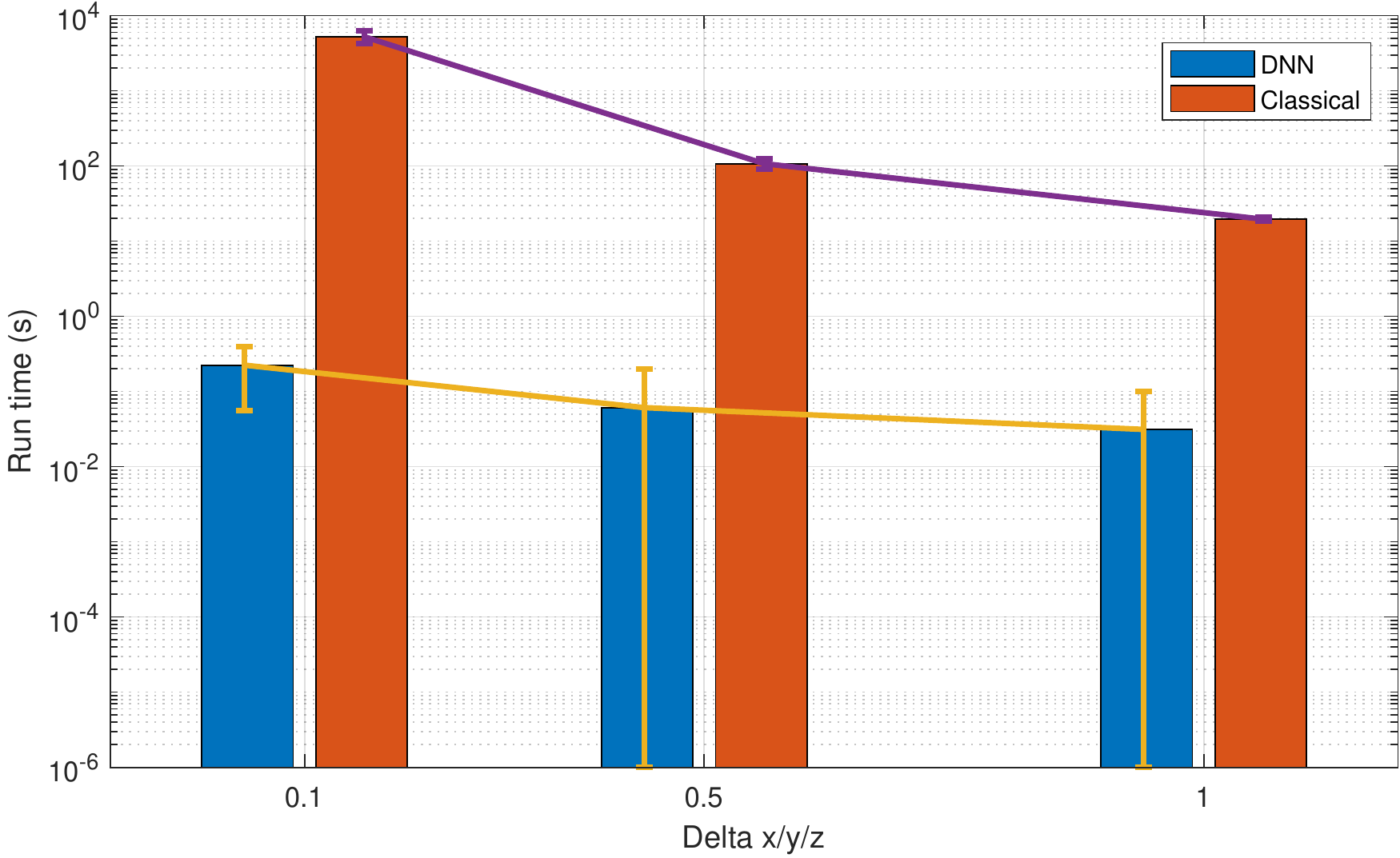}
	\caption{Run time ($\pm \sigma$) of SL model (6-DOF, $\mathbb{R}^3$) compared to the classical algorithm}
	\label{errorbar}
\end{figure}

When compared to traditional discretization methods, subspace learning algorithms show exceptionally low running time. The results from Figure \eqref{errorbar}  is recorded by the $tic$ / $toc$ functions in MATLAB\textsuperscript \textregistered \space R2017b, for implementations of both algorithms. The platform is Windows 10, and the calculations are done by a MEX worker on Intel\textsuperscript \textregistered \space Xeon\textsuperscript \textregistered \space CPU E3-1505M v5 @ 2.80GHz.

\subsection{Optimization Methods and Performance Measure}
The network in section 6.1 with layer size of $[125,100,100,125]$ is retrained with varied optimization techniques and loss measure, and their F-measures are taken from the precision and recall of the final test set ($10^3$ samples). Table (2) presents the comparison when the network is trained on a batch of $6 \times 10^4$ samples. In general, resilient backpropogation ensures better convergence given the same number of gradient updates.

\begin{table}[!h]
\label{table5}
\caption{Effects of optimization methods and performance measures on model's accuracy}
\begin{center}
\begin{tabular}{ccc}
\multicolumn{1}{c}{\bf Optimization Method}   &\multicolumn{1}{c}{\bf Performance } &\multicolumn{1}{c}{\bf F-measure ($\pm \sigma$)}
\\ \hline \\
Conj. Gradient w. P/B Restarts         &Cross entropy  & $0.3661 \pm 0.2031$\\
Variable Learning Rate G.D.         &Cross entropy  & $0.3547 \pm 0.2323$\\
One Step Secant         &Cross entropy  & $0.3574 \pm 0.2337$\\
Resilient Backpropogation         &Cross entropy  & $0.3573 \pm 0.2337$\\
Scaled Conj. Gradient         &Cross entropy  & $0.3186 \pm 0.1802$\\
Conj. Gradient w. P/B Restarts         &MSE  & $0.6686 \pm 0.3068$\\
Variable Learning Rate G.D.        &MSE  & $0.2398 \pm 0.1081$\\
One Step Secant         &MSE  & $0.6228 \pm 0.2624$\\
\textbf{Resilient Backpropogation}          &\textbf{MSE}  & $\mathbf{0.9298 \pm 0.1258}$\\
Scaled Conj. Gradient       &MSE  & $0.5938 \pm 0.3131$\\
Conj. Gradient w. P/B Restarts         &MAE  & $0.3658 \pm 0.1629$\\
Variable Learning Rate G.D.         &MAE  & $0.2256 \pm 0.1011$\\
One Step Secant         &MAE  & $0.2778 \pm 0.1178$\\
Resilient Backpropogation          &MAE  & $0.6573 \pm 0.2388$\\
Scaled Conj. Gradient        &MAE  & $0.3543 \pm 0.1366$\\
\end{tabular}
\end{center}
\end{table}

\subsection{Subspace Learning}
Finally, separate models are trained with the same architecture but various subspace size/no subspace to determine the effect of subspace embedding on convergence speed and accuracy. With a three-layer fully connected feedforward network of $80$ neurons at each layer, three models were created where the first model predicts the mapping in section 6.1, the second predicts the constant orientation workspace of manipulator at an arbitray orientation in $SO(3)$, and the third prrdicts an arbitrary 6-DOF manipulator's constant orientaiton workspace at $[0.312,5.6719,5.936]$. The last model is equivalent to no subspace embedding. All three models are tested on all test sets for subspace embedding, workspace embedding and no subspace. The results can be found in Table \eqref{table3}.


\begin{table}[!h]
\caption{Precision and recall of models ($\pm \sigma$)}
\label{table3}
\begin{center}
\begin{tabular}{ccccc}
\multicolumn{1}{c}{\bf Embedding}&\multicolumn{1}{c}{\bf Test}   &\multicolumn{1}{c}{\bf Precision } &\multicolumn{1}{c}{\bf Recall } & \multicolumn{1}{c}{\bf F-measure } 
\\ \hline \\
No subspace & n.s.& NaN & NaN & NaN\\
No subspace  &w.e.& $0.22 \pm 0.18$ & $0.35 \pm 0.13$ &$0.24 \pm 0.14$ \\
No subspace&s.e.& $0.01 \pm 0.23$ & $0.00 \pm 0.00$ &$0.01 \pm 0.01$ \\
Workspace &w.e.& $0.01 \pm 0.23$ & $0.00 \pm 0.00$ &$0.01 \pm 0.01$\\
Workspace  &s.e.& $0.85 \pm 0.17$ & $0.58 \pm 0.42$ &$0.81 \pm 0.21$\\
Workspace &n.s.& NaN & NaN & NaN\\
\textbf{Subspace} & \textbf{s.e.}&$\mathbf{0.93 \pm 0.13}$ & $\mathbf{0.76 \pm 0.37}$ & $\mathbf{0.91 \pm 0.17}$ \\
Subspace& w.e.&$0.81 \pm 0.17$ & $0.72 \pm 0.35$ &$0.82 \pm 0.18$\\
Subspace& n.s.&NaN & NaN & NaN\\
\end{tabular}
\end{center}
\end{table}

Interestingly, the subspace embedding algorithm also generalizes to arbitrary orientations in $SO(3)$. And when the task is focused on the specific subspace, subspace embedding outperforms the larger workspace embedding as well as the no subspace method. 

\subsection{Future Works}
The threshold function presented above is a naive binary classification based on one scalar value, which could potentially limit the ability of the network to generalize. Future works can thus focus on turning the classification problem into a machine learning problem, or see the workspace mapping itself as a classification rather than the regression case introduced in this work. The creation of subspace learning is mostly motivated by pragmatic concerns in the robot kinematics field, and more theoretical proofs are needed to back up the claims of improved precision and faster convergence. The mapping can also be extended to other types of workspaces, other types of robots (parallel/nonholonomic) or even be improved to include manipulability measure.

\section{Conclusion}
In this work, a modified version of subspace embedding algorithm is presented where it learns the workspace mapping of manipulators from D-H and scope parameters to a binary map. While the usage of subspace facilitates convergence, the general deep learning approach greatly reduces the computational cost by turning complex operations into basic arithmetic that took places between adjacent neurons. Subspace learning explores the feasibility of approximating inverse kinematics of robots at a three-dimensional level, comparing to the conventional attempts of converging towards a single solution; and, in a broader picture, this work opens up possibilities of reducing industrial task computational cost through methods in artificial intelligence.

\section*{Acknowledgment}
I would like to express my gratitude to Zhihong Zeng for sharing his useful insights on the paper. I sincerely thank Jiajun Mao, who has made substantial contributions in setting up a parallel computing system to generate the training samples, and configuring a Raspberry Pi 3 to implement the algorithms. I would also like to thank Bofan Xu and Matthew Austin for insightful conversations on the topic.

\printnomenclature



\bibliographystyle{IEEEtran}
\bibliography{./references}

\begin{thebibliography}{10}
\providecommand{\url}[1]{#1}
\csname url@samestyle\endcsname
\providecommand{\newblock}{\relax}
\providecommand{\bibinfo}[2]{#2}
\providecommand{\BIBentrySTDinterwordspacing}{\spaceskip=0pt\relax}
\providecommand{\BIBentryALTinterwordstretchfactor}{4}
\providecommand{\BIBentryALTinterwordspacing}{\spaceskip=\fontdimen2\font plus
\BIBentryALTinterwordstretchfactor\fontdimen3\font minus
  \fontdimen4\font\relax}
\providecommand{\BIBforeignlanguage}[2]{{%
\expandafter\ifx\csname l@#1\endcsname\relax
\typeout{** WARNING: IEEEtran.bst: No hyphenation pattern has been}%
\typeout{** loaded for the language `#1'. Using the pattern for}%
\typeout{** the default language instead.}%
\else
\language=\csname l@#1\endcsname
\fi
#2}}
\providecommand{\BIBdecl}{\relax}
\BIBdecl

\bibitem{AW14CeccarelliEclipse}
M.~Ceccarelli and E.~Ottaviano, ``A workspace evaluation of an eclipse robot,''
  \emph{Robotica}, vol.~20, no.~3, pp. 299--313, may 2002.

\bibitem{AW15LeeYoung}
\BIBentryALTinterwordspacing
T.~W. Lee and D.~C.~H. Yang, ``On the evaluation of manipulator workspace,''
  \emph{Journal of Mechanisms Transmissions and Automation in Design}, vol.
  105, no.~1, p.~70, 1983. [Online]. Available:
  \url{https://doi.org/10.1115/1.3267350}
\BIBentrySTDinterwordspacing

\bibitem{AW1BinaryMapCastelli}
\BIBentryALTinterwordspacing
G.~Castelli, E.~Ottaviano, and M.~Ceccarelli, ``A fairly general algorithm to
  evaluate workspace characteristics of serial and parallel manipulators,''
  \emph{Mechanics Based Design of Structures and Machines}, vol.~36, no.~1, pp.
  14--33, 2008. [Online]. Available:
  \url{https://doi.org/10.1080/15397730701729478}
\BIBentrySTDinterwordspacing

\bibitem{AW2WorkSpaceSE3Jin}
Y.~Jin, I.-M. Chen, and G.~Yang, ``Workspace evaluation of manipulators through
  finite-partition of se(3),'' \emph{Robotics and Computer-Integrated
  Manufacturing}, vol.~27, no.~4, pp. 850 -- 859, 2011, conference papers of
  Flexible Automation and Intelligent Manufacturing.

\bibitem{AW3MonteCarloWorkspacePeidro}
A.~Peidr\'{o}, \'{O}scar Reinoso, A.~Gil, J.~M. Mar\'{i}n, and L.~Pay\'{a},
  ``An improved monte carlo method based on gaussian growth to calculate the
  workspace of robots,'' \emph{Engineering Applications of Artificial
  Intelligence}, vol.~64, no. Supplement C, pp. 197 -- 207, 2017.

\bibitem{AW4WorkspaceParallelAnnTanase}
I.~Tanase, T.~Itul, E.~Campean, and A.~Pisla, \emph{Workspace Identification
  Using Neural Network for an Optimal Designed 2-DOF Orientation Parallel
  Device}.\hskip 1em plus 0.5em minus 0.4em\relax Dordrecht: Springer
  Netherlands, 2013, pp. 159--167.

\bibitem{AW5WorkspaceParAnnGenKuzeci}
Z.~E. Kuzeci, V.~E. Omurlu, H.~Alp, and I.~Ozkol, ``Workspace analysis of
  parallel mechanisms through neural networks and genetic algorithms,'' in
  \emph{2012 12th IEEE International Workshop on Advanced Motion Control
  (AMC)}, March 2012, pp. 1--6.

\bibitem{AW6WorkspaceParaCampean}
E.~Campean, T.~P. Itul, I.~Tanase, and A.~Pisla, ``Workspace generation for a 2
  - {DOF} parallel mechanism using neural networks,'' \emph{Applied Mechanics
  and Materials}, vol. 162, pp. 121--130, mar 2012.

\bibitem{AW7WorkspaceParaAlp}
H.~Alp, E.~Anli, and {\.{I}}.~\"{O}zkol, ``Neural network algorithm for
  workspace analysis of a parallel mechanism,'' \emph{Aircraft Engineering and
  Aerospace Technology}, vol.~79, no.~1, pp. 35--44, jan 2007.

\bibitem{AW8WorkspaceParWang}
Z.~Wang, S.~Ji, Y.~Li, and Y.~Wan, ``A unified algorithm to determine the
  reachable and dexterous workspace of parallel manipulators,'' \emph{Robotics
  and Computer-Integrated Manufacturing}, vol.~26, no.~5, pp. 454--460, oct
  2010.

\bibitem{AW9RobotWorkspaceNumCao}
\BIBentryALTinterwordspacing
Y.~Cao, K.~Lu, X.~Li, and Y.~Zang, ``Accurate numerical methods for computing
  2d and 3d robot workspace,'' \emph{International Journal of Advanced Robotic
  Systems}, vol.~8, no.~6, p.~76, 2011. [Online]. Available:
  \url{https://doi.org/10.5772/45686}
\BIBentrySTDinterwordspacing

\bibitem{AW10ParallelWrkspcOptHerrero2015}
S.~Herrero, T.~Mannheim, I.~Prause, C.~Pinto, B.~Corves, and O.~Altuzarra,
  ``Enhancing the useful workspace of a reconfigurable parallel manipulator by
  grasp point optimization,'' \emph{Robotics and Computer-Integrated
  Manufacturing}, vol.~31, pp. 51--60, feb 2015.

\bibitem{AW11SingMapParallelMacho}
E.~Macho, O.~Altuzarra, E.~Amezua, and A.~Hernandez, ``Obtaining configuration
  space and singularity maps for parallel manipulators,'' \emph{Mechanism and
  Machine Theory}, vol.~44, no.~11, pp. 2110--2125, 2009.

\bibitem{AW12HybridRobotModelPisla2013}
D.~Pisla, A.~Szilaghyi, C.~Vaida, and N.~Plitea, ``Kinematics and workspace
  modeling of a new hybrid robot used in minimally invasive surgery,''
  \emph{Robotics and Computer-Integrated Manufacturing}, vol.~29, no.~2, pp.
  463--474, apr 2013.

\bibitem{AW13NonConvexWrkspcParallelHay2002}
A.~Hay and J.~Snyman, ``The chord method for the determination of nonconvex
  workspaces of planar parallel manipulators,'' \emph{Computers {\&}
  Mathematics with Applications}, vol.~43, no. 8-9, pp. 1135--1151, apr 2002.

\bibitem{DHparameter}
R.~Hartenberg, \emph{Kinematic synthesis of linkages}.\hskip 1em plus 0.5em
  minus 0.4em\relax New York: McGraw-Hill, 1964.

\bibitem{CorkeRobotics}
P.~I. Corke, \emph{Robotics, Vision \& Control: Fundamental Algorithms in
  {MATLAB}}, 2nd~ed.\hskip 1em plus 0.5em minus 0.4em\relax Springer, 2017,
  iSBN 978-3-319-54413-7.

\bibitem{MLPApproxHornik1989}
K.~Hornik, M.~Stinchcombe, and H.~White, ``Multilayer feedforward networks are
  universal approximators,'' \emph{Neural Networks}, vol.~2, no.~5, pp.
  359--366, jan 1989.

\bibitem{MLPApproxHornik1990}
------, ``Universal approximation of an unknown mapping and its derivatives
  using multilayer feedforward networks,'' \emph{Neural Networks}, vol.~3,
  no.~5, pp. 551--560, jan 1990.

\bibitem{KuzWorkspaceGeneticPar}
\BIBentryALTinterwordspacing
Z.~E. Kuzeci, V.~E. Omurlu, H.~Alp, and I.~Ozkol, ``Workspace analysis of
  parallel mechanisms through neural networks and genetic algorithms,'' in
  \emph{2012 12th {IEEE} International Workshop on Advanced Motion Control
  ({AMC})}.\hskip 1em plus 0.5em minus 0.4em\relax {IEEE}, mar 2012. [Online].
  Available: \url{https://doi.org/10.1109/amc.2012.6197147}
\BIBentrySTDinterwordspacing

\bibitem{FengAnnSerial}
\BIBentryALTinterwordspacing
Y.~Feng, W.~Yao-nan, and Y.~Yi-min, ``Inverse kinematics solution for robot
  manipulator based on neural network under joint subspace,''
  \emph{International Journal of Computers Communications {\&} Control},
  vol.~7, no.~3, p. 459, sep 2014. [Online]. Available:
  \url{https://doi.org/10.15837/ijccc.2012.3.1387}
\BIBentrySTDinterwordspacing

\bibitem{NumericalInverseChiaverini}
S.~Chiaverini, L.~Sciavicco, and B.~Siciliano, \emph{Control of robotic systems
  through singularities}.\hskip 1em plus 0.5em minus 0.4em\relax Berlin,
  Heidelberg: Springer Berlin Heidelberg, 1991, pp. 285--295.

\bibitem{SicilianoRobotics}
B.~Siciliano and L.~Sciavicco, \emph{Robotics: Modelling, Planning and Control
  (Advanced Textbooks in Control and Signal Processing)}.\hskip 1em plus 0.5em
  minus 0.4em\relax Springer, 2008.

\bibitem{NoFreeLunchWolpert}
D.~H. Wolpert and W.~G. Macready, ``No free lunch theorems for optimization,''
  \emph{IEEE Transactions on Evolutionary Computation}, vol.~1, no.~1, pp.
  67--82, Apr 1997.

\bibitem{GoodfellowDeep}
\BIBentryALTinterwordspacing
I.~Goodfellow, Y.~Bengio, and A.~Courville, \emph{Deep Learning}.\hskip 1em
  plus 0.5em minus 0.4em\relax MIT Press, 2016. [Online]. Available:
  \url{http://www.deeplearningbook.org}
\BIBentrySTDinterwordspacing

\bibitem{RpropRiedmiller}
M.~Riedmiller and H.~Braun, ``A direct adaptive method for faster
  backpropagation learning: the rprop algorithm,'' in \emph{IEEE International
  Conference on Neural Networks}, 1993, pp. 586--591 vol.1.

\end{thebibliography}

%



\section{Appendix}
\subsection{Hint of Analogy of Loss Expectation between Subspace Average and Original Space for a Specific Case} \label{proof}

Consider first a deep neural network with any squashing activation function and an affine layer parameterized by corresponding weights and biases. The network can then be written in terms of composition of functions
\begin{equation} \label{eqNeuralNetwork}
\begin{split}
\textbf{\textit{y}}&=\underbrace{f \ldots( f(}_\text{n functions}\textbf{\textit{x}} ;\textbf{\textit{W}}^{(1)}, \textbf{\textit{b}}^{(1)}); \ldots ; \textbf{\textit{W}}^{(n)}, \textbf{\textit{b}}^{(n)})  \\
&=a(\textit{\textbf{x}};\mathit{\bm{\theta}}) 
\end{split}
\end{equation}

This model can then be used to approximate a mapping of $\mathbb{R}^n \rightarrow \mathbb{R}^m$. Referring to the maximum log likelihood estimation with $m$ training samples $L(a(\textit{\textbf{x}}^{(n)};\mathit{\bm{\theta}}),\textbf{\textit{y}}^{(n)})$ denotes the loss of a single example,

\begin{equation}
\begin{split}
&\mathbb{E}_{(\textbf{\textit{x}},\textbf{\textit{y}}) \sim p_{model}}[L(a(\textit{\textbf{x}};\mathit{\bm{\theta}}),\textbf{\textit{y}})] \\
 = &  \frac{1}{m}\sum\limits_{n=1}^m L(a(\textit{\textbf{x}}^{(n)};\mathit{\bm{\theta}}),\textbf{\textit{y}}^{(n)}) \\
 = &  \frac{1}{m}\sum\limits_{n=1}^m - \log p_{model}(\textit{\textbf{y}}|\textit{\textbf{x}};\mathit{\bm{\theta}})
\end{split}
\end{equation}

Assuming a multivariate Gaussian distribution for $p_{model}(\textit{\textbf{y}}|\textit{\textbf{x}};\mathit{\bm{\theta}})$ with identity covariance matrix,

\begin{equation}
\begin{split}
& - \log p_{model}(\textit{\textbf{y}}|\textit{\textbf{x}};\mathit{\bm{\theta}}) \\ =& - \log (\sqrt{\frac{1}{(2\pi)^m}} e^{-\frac{1}{2} \lVert \textbf{\textit{y}}- a(\textit{\textbf{x}};\mathit{\bm{\theta}})  \rVert^2 }) \\
=& -\log (\sqrt{\frac{1}{(2\pi)^m}}) + \frac{1}{2}\lVert \textbf{\textit{y}}- a(\textit{\textbf{x}};\mathit{\bm{\theta}})  \rVert^2
\end{split}
\end{equation}

Now, given $p(\textit{\textbf{x}})=\mathcal{N}(\textit{\textbf{x}}; 0_{n \times 1}, \textbf{I}_n)$, one can define a subspace as maintaining the normal nature of other elements while changing the first element to a Dirac delta distribution:

\begin{equation}
	p(x_1)=\delta(\beta)
\end{equation}

, where $\beta$ is sampled from a Gaussian distribution $p(\beta)= \mathcal{N}(\beta; 0, 1)$. Let $L(a(\textit{\textbf{x}}';\mathit{\bm{\theta}}'),\textbf{\textit{y}})$ denote the loss of an arbitrary subspace created by making an observation of the distribution $p(\beta)$, the analogy can be drawn where

\begin{equation}
\begin{split}
&\mathbb{E}_{\beta \sim p(\beta)}[L(a(\textit{\textbf{x}}';\mathit{\bm{\theta}}'),\textbf{\textit{y}})] \\
 =& - \log (\sqrt{\frac{1}{(2\pi)^m}}) + \frac{1}{2} \mathbb{E}_{\beta \sim p(\beta)}[\lVert \textbf{\textit{y}}- a([\beta,x_2',x_3' \ldots x_n']^{T};\mathit{\bm{\theta}})  \rVert^2] \\
=& - \log (\sqrt{\frac{1}{(2\pi)^m}}) + \frac{1}{2} \mathbb{E}_{x_1 \sim p(x_1)}[\lVert \textbf{\textit{y}}- a(\textit{\textbf{x}}';\mathit{\bm{\theta}}')  \rVert^2] \\
=& \mathbb{E}_{x_1 \sim p(x_1)}[L(a(\textit{\textbf{x}};\mathit{\bm{\theta}}),\textbf{\textit{y}})] \\
\end{split}
\end{equation}

This suggests a possible connection between the expectation of the loss with respect to the model/empirical distribution and the expectation of all subspace model/empirical distribution expectations.

\subsection{Time Complexity of a Classical Workspace Algorithm} \label{complexity}
According to Algorithm 2, the time complexity of $c(n)$ implemented in MATLAB\textsuperscript \textregistered \space R2017b by \cite{CorkeRobotics} can be approximated as below:

\begin{equation}\label{classicalComplexity}
\begin{split}
	&O(c(n)) \\
	\approx & O(d(f(n)))+O(no(m(n))) \\
	& +O(j(n))+O(i(m(n)) \times m^2(n)) \\
	= & O(d(m(n) \times n))+O(no(n^3))\\
	& +O(m(n) \times n)+O(i(n^3) \times n^6) \\
	= & O(d(n^4))+O(n^3)+O(n^4)+O(n^6 \times n^6) \\
	= & O(n^{12})
\end{split}
\end{equation} 

In the approximation, $O(m(n))$ indicates the time complexity of matrix multiplication, $O(i(n))$ indicates the time complexity of matrix inversion, $O(no(n))$ indicates the time complexity of computing the Euclidean norm of a vector, $O(f(n))$ indicates the time complexity of forward kinematic equation, $O(d(x))$ indicates the time complexity of the differential motion function.

The term $O(i(m(n)) \times m^2(n))$ refers to the $(\textbf{\textit{J}}^{\top}\textbf{\textit{J}} + \lambda^2 \textbf{\textit{I}})^{-1}\textbf{\textit{J}}^{\top}\textbf{\textit{e}}$ term in Algorithm 2.

Assuming elementary methods are used for matrix multiplication and $L_2$ norm calculation, while Gauss $-$ Jordan elimination method is used for matrix inversion,

\begin{equation}
\begin{split}
	m(n)=O(n^3), no(n)=O(n), i(n)=O(n^2)
\end{split}
\end{equation}

Thus, referring to Equation \eqref{classicalComplexity}, the complexity of the algorithm is

\begin{equation}
	O((c(n))^3)+O(n^3)=O(n^{36})
\end{equation}




\subsection{Numerical Inverse Kinematics} \label{ikine}
\begin{algorithmic}[1]
	\Require A manipulator object, starting joint configuration $\textbf{\textit{q}}_0$, toleration $t$, starting lambda $\lambda$, maximum rejection steps $r_{max}$, and maximum iteration steps $i_{max}$ 
	\Function{$\mathcal{K}^{-1}$}{$\xi$}
		\State $\lambda \leftarrow 0.1$ 
		\State $\textbf{\textit{e}}\leftarrow$ differential motion corresponding to $\xi$ and $\mathcal{K}(\textbf{\textit{q}}_0)$ 
		\State $\textbf{\textit{q}} \leftarrow q_i$ 
		\State $i \leftarrow 0$ 
		\State $r \leftarrow 0$ 
		
		\While{$i \leq i_{max}$}
			\State $i \leftarrow i+1$ 
	    	\If{$||\textbf{\textit{e}}|| \leq t$}
	    	\State \Return $\textbf{\textit{q}}$ 
	    	    			\EndIf
	    	\State $ \textbf{\textit{J}} \leftarrow$ manipulator Jacobian with joint angles of $\textbf{\textit{q}}$ 
	    	\State $ \textbf{\textit{dq}} \leftarrow (\textbf{\textit{J}}^{\top}\textbf{\textit{J}} + \lambda^2 \textbf{\textit{I}})^{-1}\textbf{\textit{J}}^{\top}\textbf{\textit{e}}$ 
	    	\State $\textbf{\textit{q}}_n \leftarrow \textbf{\textit{q}}+\textbf{\textit{dq}}^{\top}$ 
	    	\State $\textbf{\textit{e}}_n \leftarrow$ differential motion from $\mathcal{K}(\textbf{\textit{q}})$ to $\xi$ 
	    	\If{$||\textbf{\textit{e}}_n||<||\textbf{\textit{e}}||$}
	    		\State $\textbf{\textit{q}} \leftarrow \textbf{\textit{q}}_n$ 
	    		\State $\textbf{\textit{e}} \leftarrow \textbf{\textit{e}}_n$ 
	    		\State $ \lambda \leftarrow \frac{\lambda}{2}$ or other methods to decrease $\lambda$
	    		\State $ r \leftarrow 0$ 
   		\Else
    			\State $ \lambda \leftarrow 2\lambda$ or other methods to increase $\lambda$
    			\State $r \leftarrow r+1$ 
    			\If{$r>r_{max}$}
    			\Return $\emptyset$
    			\EndIf    			
    			\EndIf
	\EndWhile
		\State \Return $\emptyset$ 
	\EndFunction
\end{algorithmic}

\subsection{Algorithm for Experience Generation} \label{experience}

\begin{algorithmic}[1]
	\Require Number of training samples intended $i_f$ , $p($\textbf{R}$)$ , $p($\textbf{N}$)$ , where \textbf{R}$=$ $\begin{bmatrix}
	\textbf{r}_{d} \\ \textbf{r}_{a} \\ \textbf{r}_{\alpha} \end{bmatrix}$ and \textbf{N} $=$ $\begin{bmatrix}
	\textbf{n}_{x} \\ \textbf{n}_{y} \\ \textbf{n}_{z} \\ \textbf{n}_{i} \end{bmatrix}$
		\Ensure $\begin{Bmatrix} \mathbb{X},\mathbb{Y} 	
	\end{Bmatrix} =$ the training samples generated from the designed distributions

	\State $i \leftarrow 1$ 
	\State $\mathbb{X} \leftarrow \emptyset$ 
	
	\For{$i = 1,2, \ldots i_f$}
	\State (Re) initialize a manipulator object constructed from a random state of $\begin{bmatrix}
	\textbf{r}_{d} \\ \textbf{r}_{a} \\ \textbf{r}_{\alpha}
	\end{bmatrix}$ 
	\State (Re) initialize $x_{max}$ , $y_{max}$ , $z_{max}$ , $\mathcal{K}^{-1}(\xi)$ and $\mathcal{K}(\textit{\textbf{q}})$ according to Algorithm 1 \;
	(Re) initialize and populate $\fontfamily{phv}\selectfont \textbf{P}$ with steps from Algorithm 1 
	\State $\textit{\textbf{x}} \leftarrow \begin{bmatrix} \textbf{r}_{d} & \textbf{r}_{a} & \textbf{r}_{\alpha} & \textbf{n}_{x} & \textbf{n}_{y} & \textbf{n}_{z} & \textbf{n}_{i} \end{bmatrix}^\top$ 
	\State $\textit{\textbf{y}} \leftarrow $ all-zero vector with size of $x_{max}\times y_{max}\times z_{max}$ 
	\State $c \leftarrow 0$ 
	
	\For{$i=1,2, \ldots x_{max}$}
		\For{$j=1,2, \ldots y_{max}$}
			\For{$k=1,2, \ldots z_{max}$}
				\State $\textit{\textbf{y}}_{c} \leftarrow$ $\fontfamily{phv}\selectfont \textbf{P}_{i,j,k}$ 
				\State $c \leftarrow c+1$ 
			\EndFor
		\EndFor
	\EndFor
	\State $\mathbb{X} \leftarrow \mathbb{X} \cup \textit{\textbf{x}}^{(i)}$ \;
	\State $\mathbb{Y} \leftarrow \mathbb{Y} \cup \textit{\textbf{y}}^{(i)}$ \;	
	\EndFor
\end{algorithmic}

\end{document}